\documentclass[journal]{IEEEtran}
\usepackage{latexsym,bm,amsmath,amssymb}
\usepackage{enumerate}
\usepackage{graphicx}
\usepackage{epstopdf}
\usepackage{subfigure}
\usepackage[ruled,vlined]{algorithm2e}
\usepackage{algorithmic}
\usepackage{amsfonts,amssymb,amsmath}
\usepackage{dsfont}
\usepackage{multirow}
\usepackage[table]{xcolor}
\usepackage{colortbl}
\usepackage[T1]{fontenc}
\usepackage{bm}
\usepackage{setspace}
\usepackage{color}
\usepackage{cite}

\graphicspath{{figures/}}

\begin{document}

\title{A Novel Biologically Mechanism-Based Visual Cognition Model \\
--\begin{LARGE}Automatic Extraction of Semantics, Formation of Integrated Concepts and Re-selection Features for Ambiguity\end{LARGE}
}


\author{Peijie Yin, Hong Qiao,~\IEEEmembership{Senior~Member,~IEEE,}, Wei Wu, Lu Qi, YinLin Li, Shanlin Zhong, Bo Zhang
\thanks{H. Qiao, W. Wu, L. Qi, Y. Li, S. Zhong are with the State Key Lab of Management and Control
for Complex Systems,Institute of Automation, Chinese Academy of Sciences,
Beijing 100190, China (e-mail: hong.qiao@ia.ac.cn).}
\thanks{P. Yin, B. Zhang are with the Institute of Applied Mathematics, Academy of Mathematics and Systems Science, Chinese Academy of Science.}
\thanks{ This work was supported by the National Science Foundation of China under grant 61210009, and the Strategic Priority Research Program of the CAS (grant XDB02080003).}}


\maketitle
\IEEEpeerreviewmaketitle

\begin{abstract}
Integration between biology and information science benefits both fields. Many related models have been proposed, such as computational visual cognition models, computational motor control models, integrations of both and so on. In general, the robustness and precision of recognition is one of the key problems for object recognition models.

In this paper, inspired by features of human recognition process and their biological mechanisms, a new integrated and dynamic framework is proposed to mimic the semantic extraction, concept formation and feature re-selection in human visual processing. The main contributions of the proposed model are as follows:

\begin{enumerate}[(1)]
\item
\emph{Semantic feature extraction}: Local semantic features are learnt from episodic features that are extracted from raw images through a deep neural network;

\item
\emph{Integrated concept formation}: Concepts are formed with local semantic information and structural information learnt through network.

\item
\emph{Feature re-selection}: When ambiguity is detected during recognition process, distinctive features according to the difference between ambiguous candidates are re-selected for recognition.

\end{enumerate}
Experimental results on hand-written digits and facial shape dataset show that, compared with other methods, the new proposed model exhibits higher robustness and precision for visual recognition, especially in the condition when input samples are smantic ambiguous. Meanwhile, the introduced biological mechanisms further strengthen the interaction between neuroscience and information science.
\end{abstract}


\section{Introduction}

\IEEEPARstart{W}{ith} the integration of neuroscience and information science, more and more biological mechanisms have been applied in computational models, which promotes the development of biologically inspired models. On the one hand, inspired by recent findings in biology, these models outperform classic algorithms in performance and efficiency. On the other hand, related neural mechanisms, which are introduced into computational models, could be validated and testified to promote the development in neuroscience.

Vision is one of the key interdisciplinary research directions between neuroscience and information science. Mechanisms of primate and human in visual processing and cognition have been introduced into computational cognition model.

HMAX model tries to mimic the functions of primate visual system layer by layer \cite{riesenhuber1999hierarchical}. 

The main difference between HMAX and other hierarchical architectures (such as hand-crafted hierarchical features \cite{krizhevsky2012imagenet}, convolutional neural networks \cite{guo2007primal}, and etc.) is that it focused on reproducing anatomical, physiological and psychophysical properties of the ventral pathway of visual system \cite{dura2012top}, which consists of $V1$, $V2$, $V4$ and inferior temporal (IT) cortical areas.

After its first publication in 1999, this well-known model has been further developed and improved in different aspects \cite{serre2004realistic,irshad2013automated,theriault2013extended,hu2014sparsity}. For example, many researchers modified the original HMAX model by adding feedback process to improve the recognition precision \cite{li2015enhanced,liu2015hmax}. 

Other modifications include adding sparsity to the convolutional layer \cite{hu2014sparsity}, enhancing the architecture by adding specific layers to the model \cite{theriault2013extended,han2010biologically}, and changing strategies of feature selection and filtering properties \cite{mishra2010hierarchical}. Frontier researchers try to introduce mechanisms of attention into visual model. Itti proposed a saliency-based model based on the saliency map theory in human visual system \cite{itti1998model} and combined attention with object recognition \cite{miau2001neural,walther2002attentional}. Spatial information of an object is introduced by modeling the dorsal pathway in vision system. It has been implement by Bayesian inference \cite{chikkerur2010and} and saliency model \cite{borji2014look}. Mechanisms in middle and high level cortices are also a hot topic in the area. Based on HMAX and deep neural network, Qiao et al. developed a series of models introducing association \cite{qiao2014introducing}, attention\cite{qiao2014biologically} to the model. The introduced mechanisms show good performance on object classification and identification tasks.

Robustness, i.e., the ability of generalization is one of the key objectives and motivations in these visual cognition models. However, recent findings \cite{nguyen2014deep,goodfellow2014explaining} point out that even the state-of-art deep hierarchical networks suffer from tiny disturbance and transformation. It is shown that tiny perturbation may cause significant difference in the output of hierarchical network models \cite{goodfellow2014explaining}. 

However, human has extraordinary ability to deal with difficult object recognition task with various viewpoints, scales, deformation, and ambiguity. According to biological findings, objection recognition tasks involve multiple cortices and many sophisticated mechanisms including preliminary cognition, top-down attention \cite{desimone1995neural}, semantic and conceptual memory \cite{jhuang2007biologically,kintsch2014representation,patterson2007you}. Lake, Salakhutdinov and Tenenbaum recently \cite{lake2015human} employs semantics and concepts explicitly and achieves significant improvement in robustness of one-shot character recognition.

In this paper, we build a Biological mechanism based Semantic Neural Network model (BSNN), which extracts semantic information hierarchically and forms concepts with corresponding probabilities. The model is trained sequentially and generate hierarchical information layer by layer. 

To mimic the biological mechanisms, the model firstly trains a neural network to extract episodic features; then it integrates the learnt episodic features into the semantic features. To encode the structure information, the model learns structural relationships between semantics and represents them as population vectors. With the population vectors, concepts for categories are formed in a probabilistic way. The proposed model also applies two dynamic updating strategies, feature re-selection for adaption to ambiguous condition, and online training for new concepts. By mimicking and implementing neural mechanisms in visual processing, the model achieves robustness to various ambiguous images with small training samples. It is also more efficient to diminish uncertainty by semantics and concepts with the ability of generalization.

The rest of this paper is organized as follows. Section II introduces biological evidence of the proposed model. Section III explains the framework and methods in the BSNN. Section IV presents how the experiments are conducted and shows experimental results. Section V summarizes current work and points out future direction.

\section{Biological Evidence}
In this paper, several biological mechanisms are introduced into the new framework to mimic semantic extraction, concept formation and feature re-selection process in human visual processing. Here, related biological evidence has been reviewed and discussed for its validity of later implementation.

\subsection{Semantic Feature Extraction}
Two different types of memory are stored in the brain: episodic memory and semantic memory \cite{patterson2007you,tulving1985many}. Episodic memory stores events and detailed contextual information, while semantic memory extracts regularities from different spatial-temporal events and forms perceptual categories, complex concepts and relations \cite{tulving1985many}. This requires that extraction of regularities or semantics should be carried out over episodes \cite{moscovitch2005functional,sweegers2014neural}. Since hippocampus is involved with storage of episodic memory and prefrontal cortex contributes to organization of information, the extraction process could be achieved via hippocampus and mPFC (medial prefrontal cortex) interaction \cite{benchenane2010coherent,kumaran2009tracking,van2010persistent}. The extracted semantic information could be used for later tasks.
\subsection{Structural Information}
It has been proposed that objects could be described with parts and their positional and connectional relationships \cite{biederman1987recognition,marr1978representation}. For example, neurons in V4 are tuned for contour fragment orientation with specific object-relative position \cite{pasupathy2001shape}. In other words, one V4 neuron could respond to convex curvature at bottom right (such as 'b'), but not to that at top right (such as 'p'). Thus, in V4 area, neurons respond to individual contour fragments and their relationships are encoded in population responses \cite{pasupathy2002population}. In PIT (posterior inferior temporal cortex), neurons integrate information on multiple fragments \cite{pasupathy2001shape}. Thus, the integrated explicit representations of multi-part configurations could be encoded in IT.

\subsection{Selective Attention}
Attention is required when people carry out various tasks, since relevant environmental stimuli and information should be selected and processed in the brain \cite{desimone1995neural,petersen2012attention}. Several brain areas are activated in the attention process, such as frontal eye fields (FEF), anterior cingulate, frontal cortex, and etc. \cite{green2008using}. Visual attention usually consists of active exploration of the environment, selection of task-related information and suppression of distraction. When visual stimuli is not clear for the task, visual attention process could suppress distraction from location of previous attention focus and find new positions for the search of related information \cite{golomb2008native}.

\section{The Framework}
In this section, we present the structure of the proposed framework. Firstly, the outline of the framework is described. Secondly, the algorithms of semantic feature extraction, integrated concept formation and feature re-selection are given in details. 

\subsection{Outline of the Framework}
\begin{figure*}[!htb]
\centering
\includegraphics[width=1.0\textwidth]{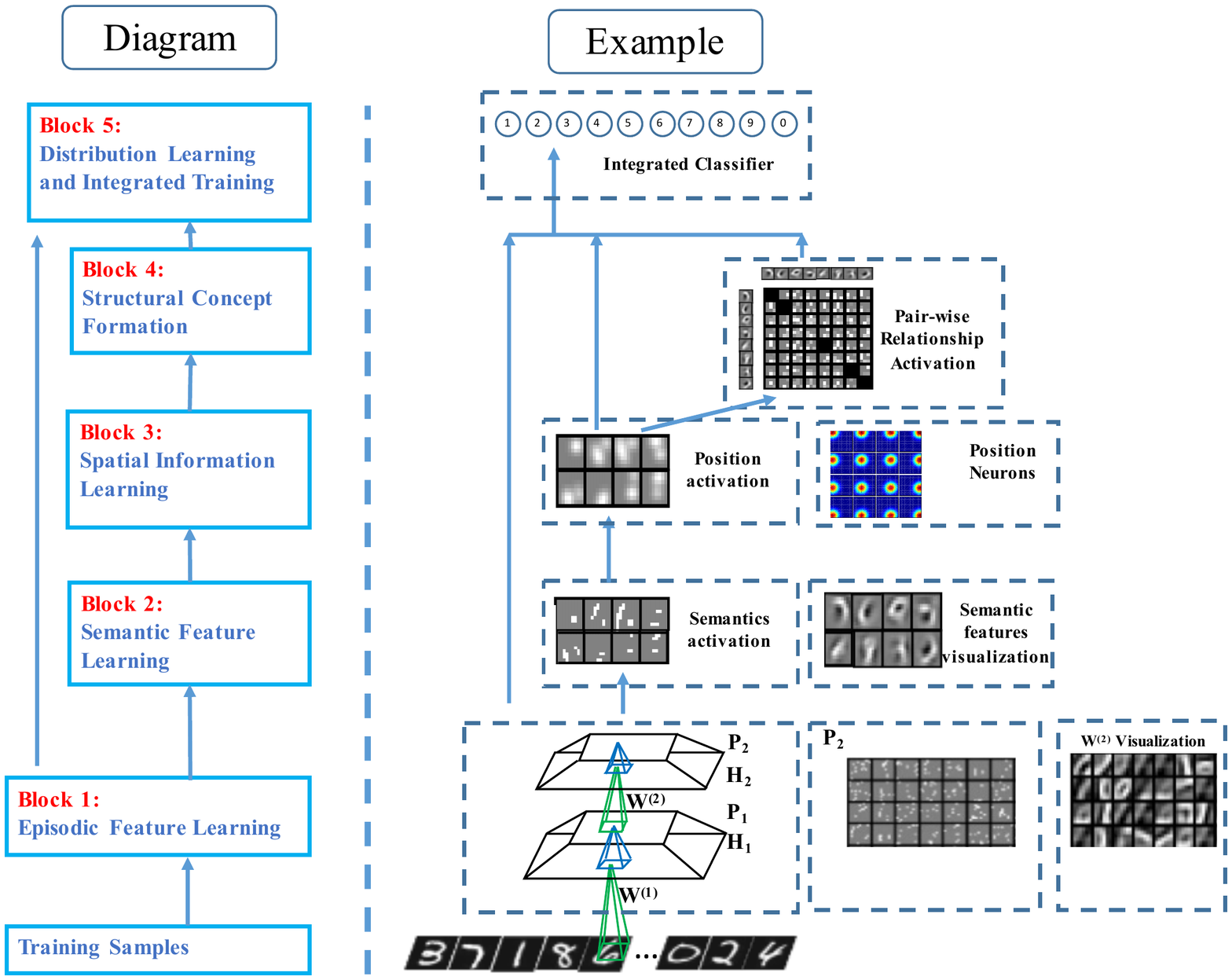}
\caption{Training procedure. On the left is the diagram of the model. On the right is an illustrative example of training one image and the visualization of the intermediate results.}
\label{fig:training_procedure}
\end{figure*}

\begin{figure*}[!htb]
\centering
\includegraphics[width=1.0\textwidth]{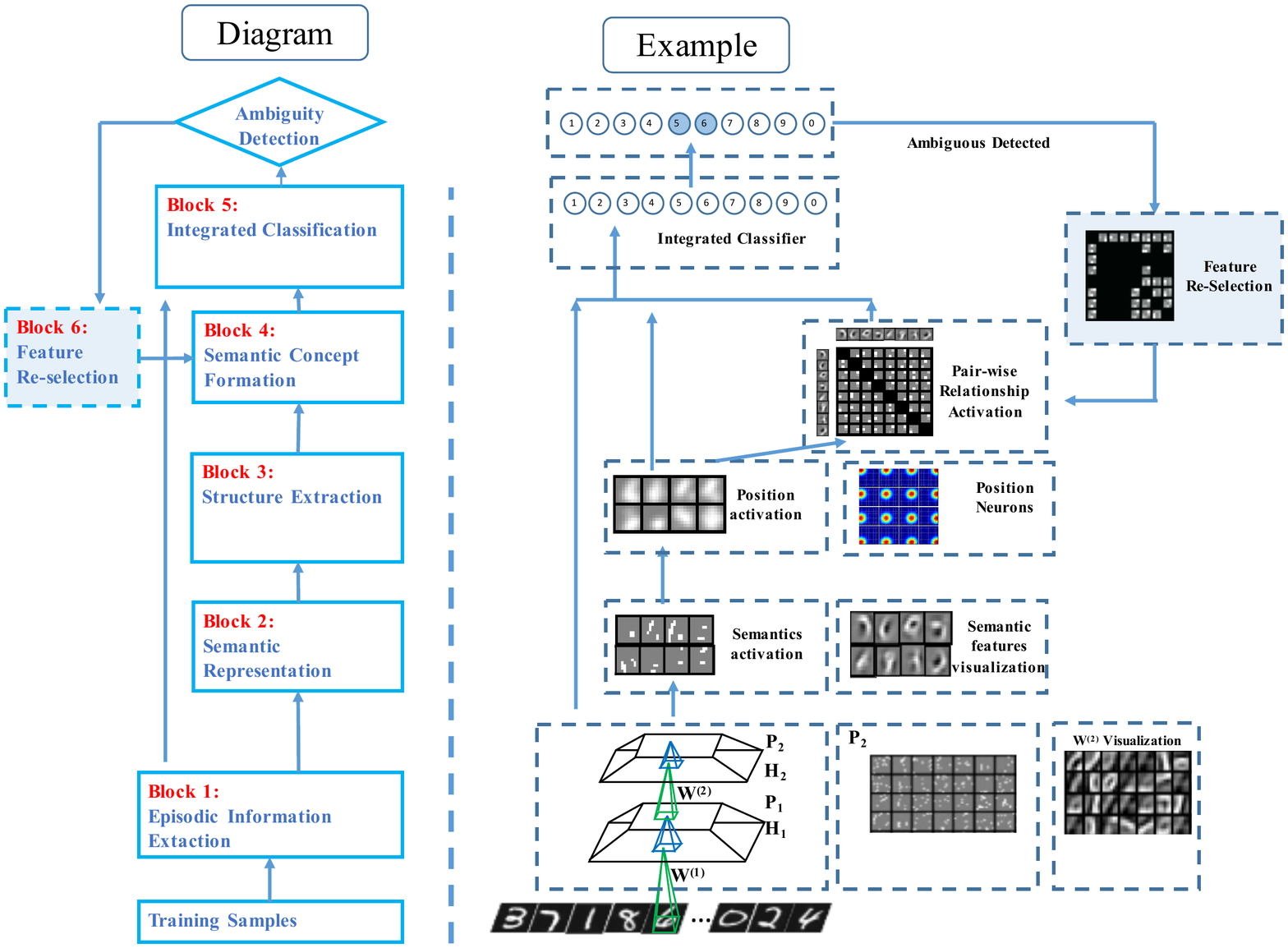}
\caption{Recognition procedure. On the left is the diagram of the recognition process. On the right is the example.}
\label{fig:Recognition_procedure}
\end{figure*}

Figure \ref{fig:training_procedure} shows the training procedure of the model. Figure \ref{fig:Recognition_procedure} shows the recognition and online updating procedure.

\subsubsection{Block1}
Primary episodic feature extraction block, where episodic features are extracted from the original image directly. Block 1 includes a four-layer convolutional deep belief network (CDBN). The network is trained layer by layer without supervision. The output of the layer is the activation states of the last layer (for block $2$) and the learnt connection  weights (for block $3$).
\subsubsection{Block2}
Semantic feature extraction block, which extracts semantic features from the learnt episodic features in Block $1$. 

A cluster-based method is applied here, which provides a better abstract description of the object than the feature maps.

\subsubsection{Block3}
Semantic spatial information learning block, where spatial positions of semantic features are learnt based on the output of Block 1 and Block 2. Spatial information is encoded based on position-related population neurons.
\subsubsection{Block4}
Structural concept formation block, where relationships between semantic features are formed from the spatial information in Block 3. Relationships are encoded via orientation-related population neurons. For each input sample, a relationship matrix is generated to represent the global structure.
\subsubsection{Block5}
Integrated recognition block, which combines episodic and semantic features together for recognition. Block $5$ in Fig. $1$ learns weights of episodic and semantic features from different pathways, which uses for integrated classification in Block $5$ of Fig. $2$.
\subsubsection{Block6}
Feature re-selection block, which copes with ambiguous situations dynamically. During the training procedure in Fig. 1, the correlation between extracted features and categories will be learnt. When there are two or more candidate recognition results, features that are more discriminative between candidates will be selected for further classification.

\subsection{Episodic Features Learning with Unsupervised Deep Neural Network (Block 1)}
In human, episodic memory is the memory that represents experiences and specific events in time, from which people can reconstruct the actual events that took place at any given time \cite{tulving1985many}. Episodic memory is one of the basic forms of explicit memory and considered as the source of other forms of memory \cite{tulving2002episodic}.

\begin{figure}[!htb]
	\centering
	\includegraphics[width=0.8\linewidth]{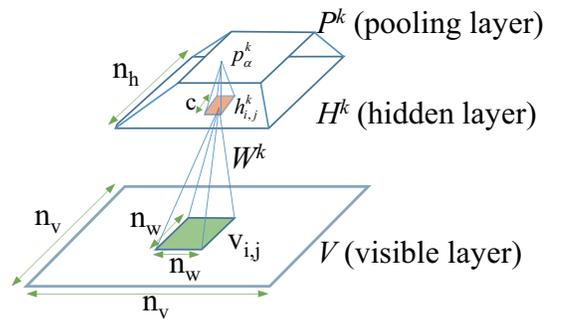}
	\caption{A CRBM model with probabilistic max-pooling. Only the kth
		channels of $\mathbf{H}$ and $\mathbf{P}$ are shown for simplicity.}
	\label{fig:crbm}
\end{figure}

In this paper, episodic features are extracted via an unsupervised deep neural network. Unsupervised convolutional deep belief network (CDBN) is first introduced in Ng's work \cite{lee2009convolutional} for feature extraction tasks. In our previous work \cite{qiao2015biologically}, CDBN has been used to extract episodic information from the image. As an unsupervised model, CDBN is able to extract good local features and encode common components by minimizing the reconstruction error, which ensures good performance in recognition. CDBN is composed of stacked Convolutional Restricted Boltzmann Machine (CRBM). CRBM, as a variant of RBM, can infer the original input from the activation and minimize the reconstruction error. Thus, the visual information could be retrieved from memory, which is similar to human. The structure of CRBM is showed in Fig \ref{fig:crbm}.

As shown in Fig. \ref{fig:crbm}, each CRBM includes three layers: visible layer $\mathbf{V}$, hidden layer $\mathbf{H}$ and pooling layer $\mathbf{P}$. $n_v$ and $n_h$ are the widths of $\mathbf{V}$ and $\mathbf{H}$, respectively. $\mathbf{H}$ has $K$ groups of feature maps, which is denoted by $\mathbf{H}^k~(k=1,2,\cdots,K)$. $\mathbf{H}$ is connected with $\mathbf{V}$ by the shared local weights $W^k$ with width $n_w$. So the width of $\mathbf{H}^k$ is calculated as $n_h = n_v - n_w + 1$. Let $v_{i,j}^k$ represent an unit in layer $\mathbf{V}$ with row index $i$ and column index $j$, 
and $h_{i,j}^k$ stands for an unit in layer $\mathbf{H}^k$. Layer $\mathbf{P}$ is the pooling layer of $\mathbf{H}$. The unit $p^k_\alpha$ is obtained by pooling from a specific $c\times c$ block in $\mathbf{H}^k$ denoted by $B_\alpha$. So $\mathbf{P}$ also has $K$ groups of feature maps $P^k~(k=1,2,\cdots,K)$ with width $n_p=n_h/c$. In mathematics, the CRBM  model is a special type of energy-based models. Given inputs $\mathbf{V}$ and hidden layer $\mathbf{H}$ with binary feature maps $\mathbf{H}^k$, the energy of each possible state ($v$, $h$), where $v\in \mathbb{R}^{n_v \times n_v}$ and $h\in \mathbb{B}^{n_h \times n_h \times K}~(\mathbb{B}=\{0,1\})$, is defined in (1)

\begin{eqnarray}
E(v,h) &=& -\sum_{k=1}^{K}\sum_{i,j=1}^{n_h}h_{i,j}^k (\tilde{W}^k * v)_{i,j} -\sum_{k=1}^{K}b_k \sum_{i,j=1}^{n_h} h_{i,j}^k  \nonumber  \\
                             & &   - a\sum_{i,j=1}^{n_v}v_{i,j} + \frac{1}{2}\sum_{i,j=1}^{n_v} v_{i,j}^2, \label{eq:energy}
\end{eqnarray}

where $h_{i,j}^k$ meets the constraint

\begin{eqnarray}
 \sum_{(i,j)\in B_{\alpha}} h_{i,j}^k \leq 1, \forall k,\alpha.
\end{eqnarray}
Here, $\tilde{W}^k$, representing the $180$-degree rotation of matrix $W^k$, is the convolutional kernel, $*$ denotes the convolution operation, $b_k$ is the shared basis of all units in $\mathbf{H}^k$, and $a$ is the shared basis of visible layer units.

The CRBM can be trained with Contrastive Divergence (CD), which is an approximate Maximum-Likelihood learning algorithm \cite{hinton2002training}.  By training CRBMs sequentially, the CDBN is also trained.

However, since it is trained without supervision, the features may be not distinctive enough for classification. Meanwhile, useful features sometimes could shrink to a small set and thus the generalization ability is limited. Moreover, because it focuses on minimization of the reconstruction error, it could not go deeper to the semantic level and extract the structure information. To overcome this drawback, this paper introduces extraction of semantic features to enhance the ability of distinction of features.

As a generative model, CDBN also has the ability to achieve reconstruction from activation. So the model can recall the original input image by reconstruction as augmentation of training data.

In this paper, we train a two-layer CDBN and apply it to extract features from original images. We also reconstruct the input images from the activation of the output layer for visualization and data augmentation when the number of training samples is relatively small. The visual reconstruction $v'$ is defined by
\begin{eqnarray}
v' = \sum\limits_k {{s_k}({W^k}*{H^k})}
\end{eqnarray}
 where $s_k$ denotes the weights of the connection between $W^k$ and $H^k$. $s_k$ is learned by feature re-selection, which is described in detail in Block $6$.

\subsection{Semantic Representation Learning based on Episodic Features (Block 2)}
Semantics has multiple definitions in different fields, such as linguistics \cite{berg2002semantics,longobardi2001comparative}, cognitive science \cite{mitchell2010composition,binder2011neurobiology}, artificial intelligence \cite{eiter2008combining,aksoy2011learning} and etc. In cognitive science, semantic memory is about facts that capture the internal properties about an object \cite{tulving1985many,binder2011neurobiology}. Human use semantic memory to store the category and abstract information about the object and distinguish one category of objects from others. Binder and Desai \cite{binder2011neurobiology} proposed that modality-specific semantic memory is encoded in the corresponding cortex. Convergence of these findings, semantic information in vision is represented in the similar form of visual episodic features, but more abstract and discriminative.

Inspired by the above mentioned properties of semantic memory in neuroscience, a reasonable hypothesis is that semantic features for visual task are formed based on those learnt hierachical episodic features. Semantic features are more likely to activated by diverse properties of an object. For visual recognition, each property, like a stroke or a shape, represents a general cluster of patches rather than a certain patch. In this way, a formalized description of semantic features is given as follows:

Denote the reconstruction function $ f_{recon}:\mathbb{F}\rightarrow\mathbb{I}$, where $\mathbb{F}$ is the space of episodic features, $\mathbb{I}$ is the space of input images.  

Given patches $\mathbf{V}' = \{ {v_1'},{v_2'},...,{v_n'}\} $  (${v_i'}$ is reconstructed from the $i$th episodic feature, where $1 \le i \le n$ and $n$ is the number of episodic features), find $K$ groups in $\mathbf{V}'$ based on similarities of patches ($K$ is a relative small number compared to the size of $\mathbf{V}$). Divide corresponding episodic features into groups according to their reconstructions. For each group of patches, find $v^j$ $(v^j\in f_{recon}(\mathbb{F}),j=1,..,K)$ as a representative of the group which minimizes the loss of imformation. Representatives denoted as $\{S_j,j=1,..,K\}$, $\{S_j\}$ are semantic features abstracted from previous episodic features.

The objective of semantic clustering is to find:
\begin{eqnarray}
\mathop {\arg \min }\limits_S \sum\limits_{j = 1}^K {{{\sum\limits_{{v_i'} \in {S_j}} {\left\| {{v_i'} - {S_j}} \right\|} }^2}}
\end{eqnarray}

where $\left\|\cdot\right\|$ refers to the metric from $I$ restricted to $f_{recon}(\mathbb{F})$.

In our model, for computational convenience, we use k-means and $L2$ metrics to iteratively find the desired ${S_i}$. Given an initial set of k-means $\{ m_1^{(1)},...,m_k^{(1)}\} $, the algorithm proceeds by alternating between two steps:

Assignment step: Assign each observation to the cluster whose mean yields the least within-cluster sum of squares (WCSS).
\begin{eqnarray}
S_j^{(t)} = \{ {v_i'}:{\left\| {{v_i'} - m_j^{(t)}} \right\|^2} \le {\left\| {{v_i'} - m_l^{(t)}} \right\|^2}\  \forall l,1 \le l \le k\}
\end{eqnarray}
where each ${v_i'}$ is assigned to exactly one ${S_{(t)}}$, even if it could be assigned to two or more of them.

Update step: Calculate the new means to be the centroids of the observations in the new clusters.
\begin{eqnarray}
m_j^{(t + 1)} = \frac{1}{{\left| {S_j^{(t)}} \right|}}\sum\limits_{{v_i'} \in S_j^{(t)}} {{v_i'}}
\end{eqnarray}

where $\left| {S_j^{(t)}} \right|$ denotes the number of elements in $S_j^{(t)}$.

With a few iterations as mentioned above, the patches could form different clusters. We use the center of clusters as semantic features.

\subsection{Structural Learning with Population Coding (Block 3 \& Block 4)}
In the context of object recognition, spatial structure information is highly valuable but difficult to find a proper representation. Neuroscience researches \cite{stokes2009top} reveal that human brain processes this kind of information with a population of neurons. Taking the population of neurons related to orientation as an example, each neuron has a preferred direction; the closer to the preferred direction, the more a direction of stimulus will activate a certain neuron. The relationship between preferred stimulus and the rate of activation can be represented as a Gaussian-like curve. With many populations together, not only the input stimulus but also the uncertainty of the stimulus could be encoded by the rate of activations among the population of neurons.

In this paper, we define two kinds of structural features, (1) position features, the relative position of a component to the object center (2) relationship features, a relative structure which consists of spatial directions and distances between semantic components. The former feature captures the spatial positions of different semantic features in an input sample. The latter feature represent global concepts of how different features are organized together.

Inspired by the population coding mechanism in biological neural system, the structure is encoded by two population of neurons, one population encodes the positions whereas the other population encodes the relative relationship between different components. In this paper, the two populations of neurons are denoted as position neurons PNeurons and the relationship neurons RNeurons.

\begin{figure}[!htb]
	\centering
	\includegraphics[width=0.7\linewidth]{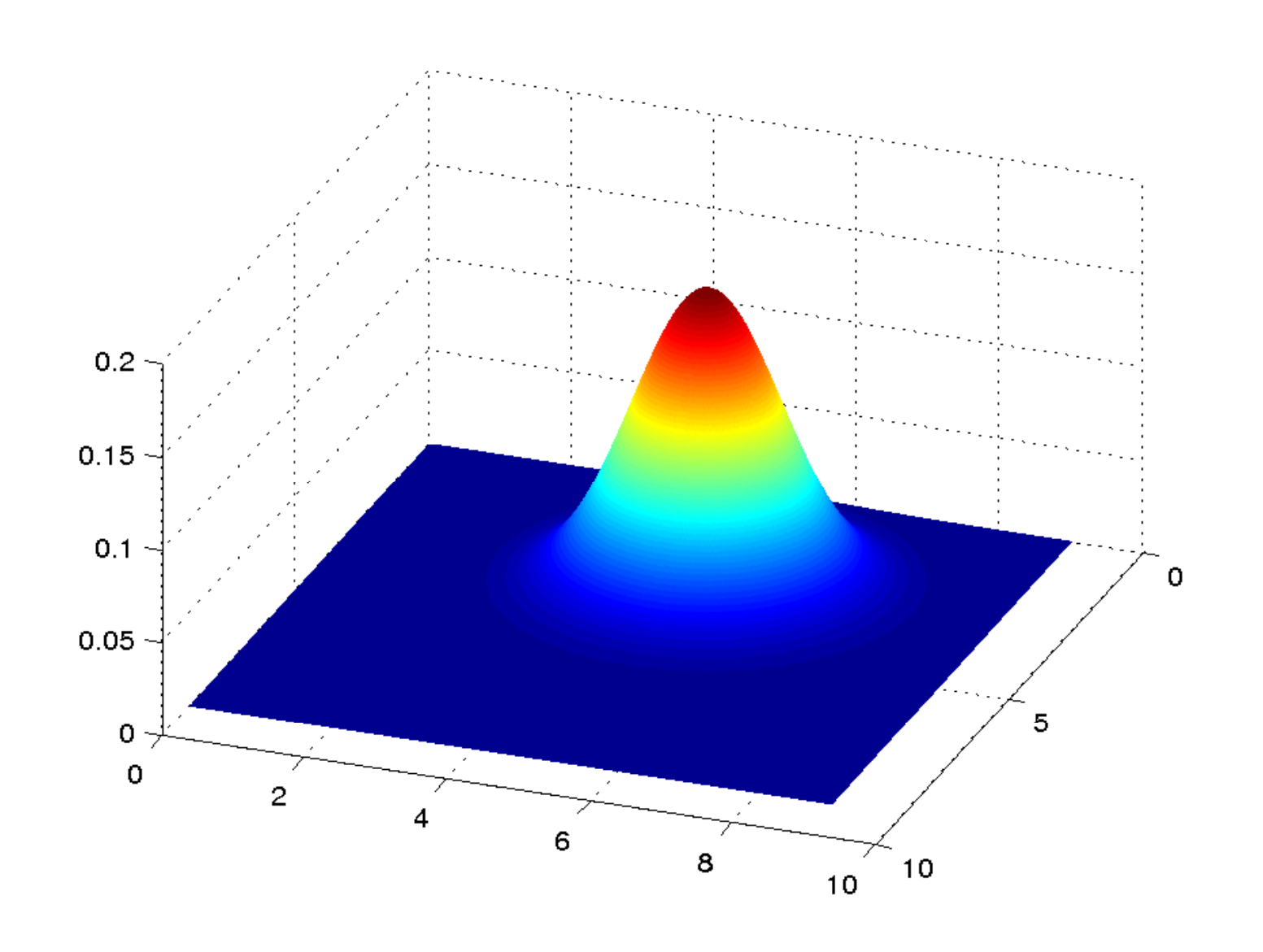}
	\caption{Illustration of the Gaussian tunning function of a PNeuron. The neuron is sensitive to semantic feature inputs near the center.}
	\label{fig:single_PNeuron}
\end{figure}

\begin{figure}[!htb]
	\centering
	\includegraphics[width=0.65\linewidth]{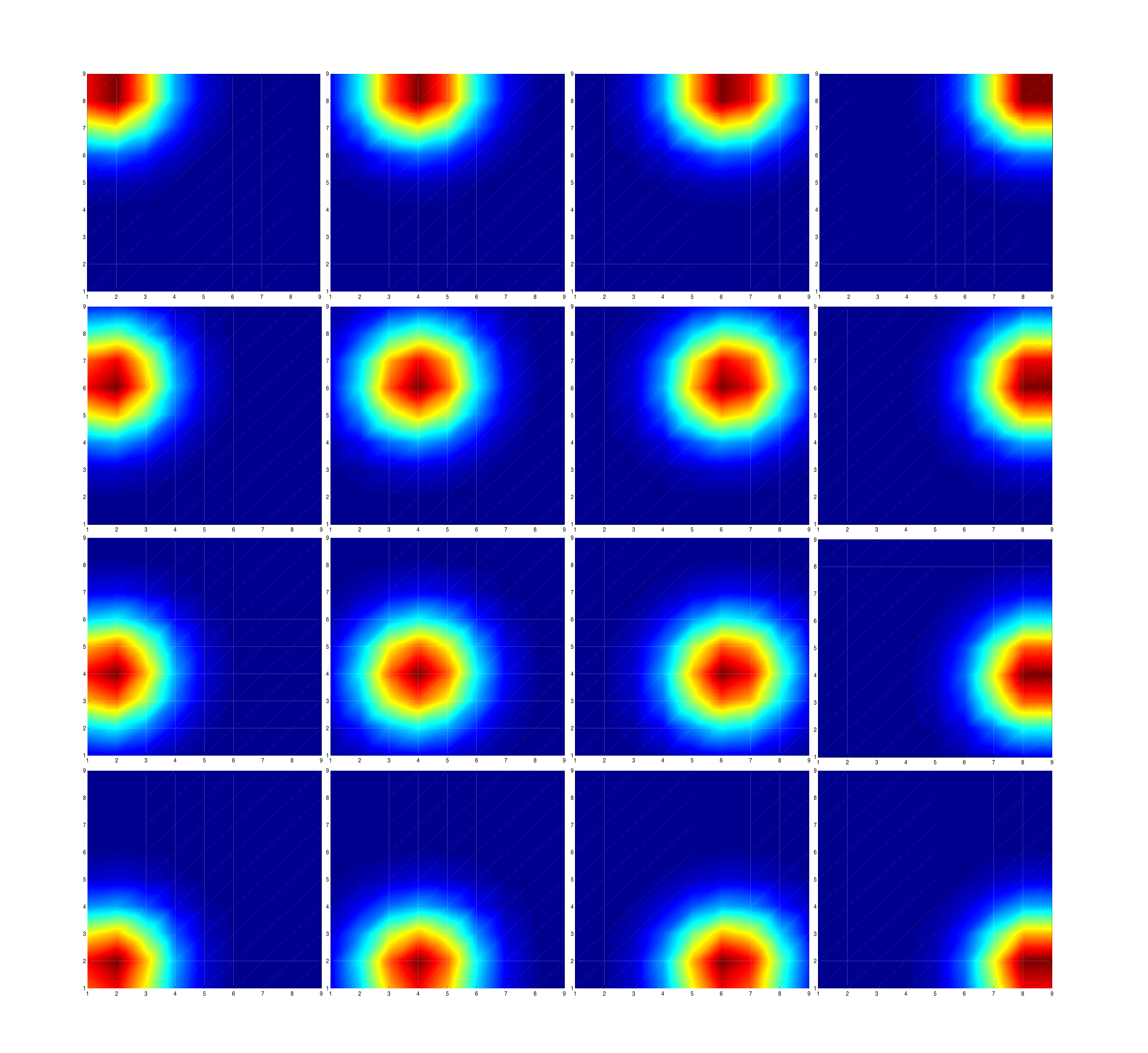}
	
	\caption{Tunning functions of uniformly distributed PNeurons. Each of square represent a PNeuron's tuning function. The outputs of these 16 PNeurons form a 4$\times$4 position matrix (PM).}
	\label{fig:multiple_PNeuron}
\end{figure}

Fig. \ref{fig:single_PNeuron} and Fig. \ref{fig:multiple_PNeuron} give examples of PNeurons. A single neuron is like a gaussian filter (Fig. \ref{fig:single_PNeuron}), in consistent with neuroscience researches. Each neuron has its own preferred position, the position that mostly activates the neuron. With multiple PNeurons together (e.g. 16 PNeurons as shown in Fig. \ref{fig:multiple_PNeuron}), the population will output spatial representations of semantic features. 

\begin{figure}[!htb]
\centering
\includegraphics[width=1.0\linewidth]{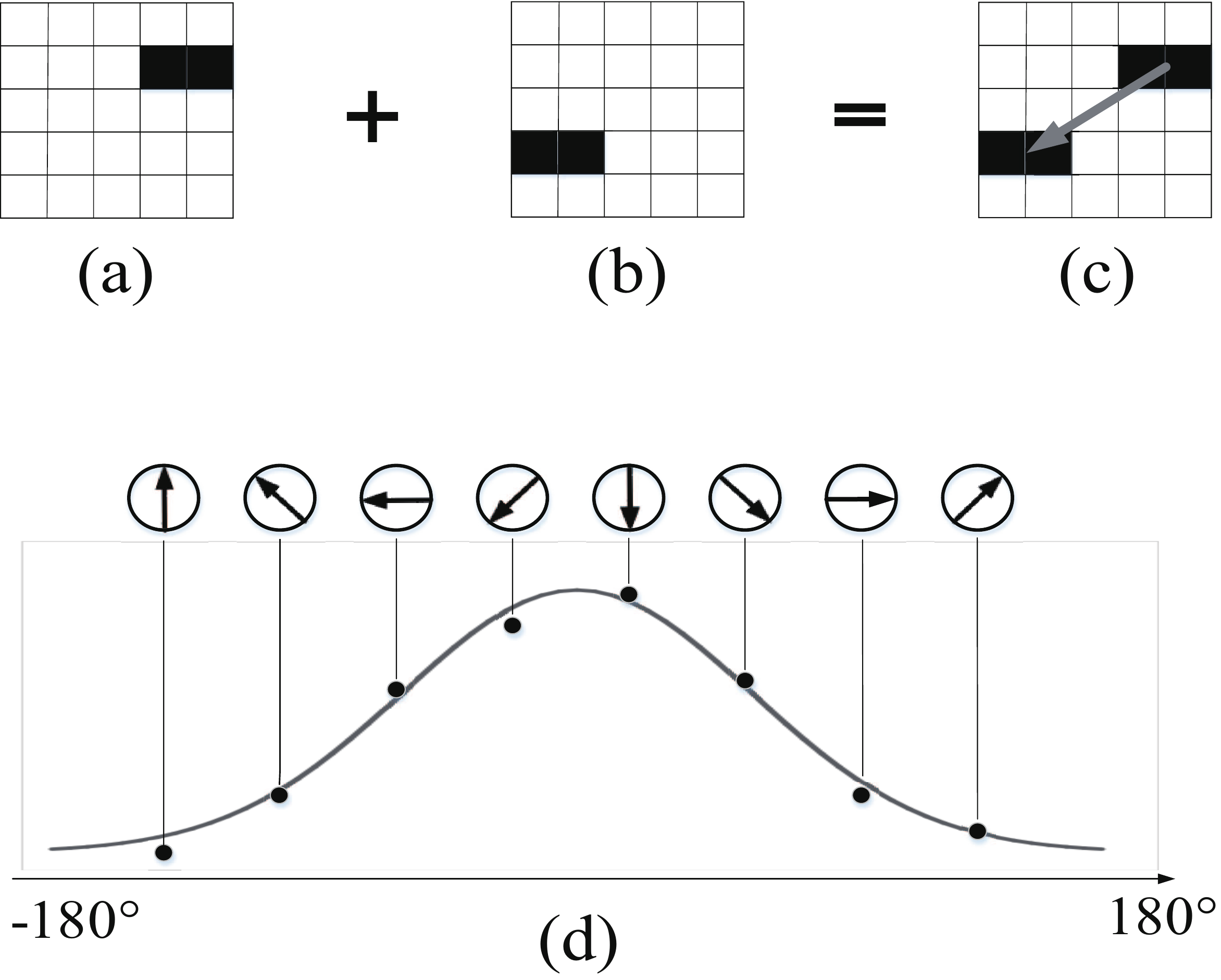}
\caption{Example of representing direction via population coding. (a) and (b) are activation and visualization of two semantic features, (c) generate the relative direction between different semantic features. (d) is the output of relationship neurons, encoding the direction.}
\label{fig:population_coding}
\end{figure}

The activation process is almost the same with RNeurons' except RNeurons prefer orientations rather than positions. Fig. \ref{fig:population_coding} shows the detailed process of how RNeurons are activated. Each neuron has a preferred direction that maximizes its activation. For a certain neuron, the activation responding to a direction is characterized as a Gaussian, depending on the difference between the input and the preferred direction. 

From the output of semantic layer (semantic feature maps), one can locate the position of each semantic feature via population of PNeurons. The tuning function of a PNeuron is a 2D Gaussian function centered in a certain position, which represents its probability of activation with different input position. With multiple PNeurons, each semantic feature map could be encoded as a new map about where the semantic features are most likely located in the image. In this paper, for computational consideration, the center position of PNeurons are uniformly distributed over the map and the Gaussian functions are discretization with the same size of semantic feature maps. For a input feature map, the aggregated output of all PNeurons forms a matrix according to their center position, (denoted as position matrix or PM in the rest of paper).

\begin{figure}[!htb]
\centering
\includegraphics[width=1.0\linewidth]{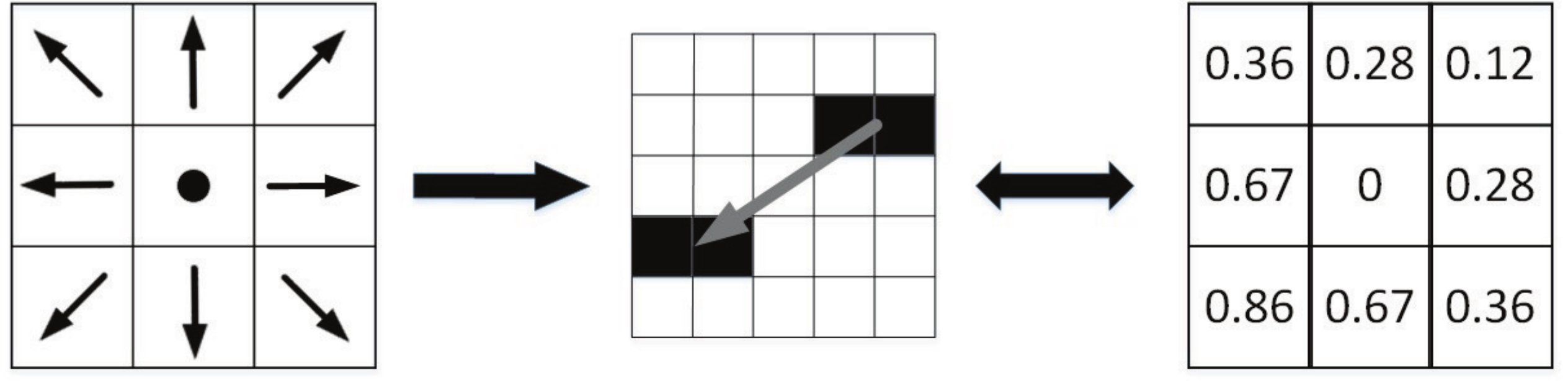}
\caption{Illustration of relationship matrix with 8 RNeurons. Figure on the left is an example of relationship neurons for certain two features. Each surrounding neuron has its own preferred direction; the center neuron encodes the distance. On the right is the actual activation of the left neurons, when two input features with a spatial relationship like the middle one.}

\label{fig:relationship_matrix}
\end{figure}

With the semantic features and their positions (by sampling from the PM), the population of RNeurons will output the relative relationship between different features. For every two semantic features, the paper defines a relationship matrix to describe their position relationships. As is shown in Fig. \ref{fig:relationship_matrix} with 8 RNeurons, for an input direction, the output of all the RNeurons can be represented as a feature in the relationship matrix. The center node of the matrix is to encode the distance between the input position. 

The output of relationship neurons is structural semantic information of input sample. As is shown above, structural information is distributed encoded by a population of neurons. Each neuron responds to two specific semantic features and one preferred direction. Thus, encoded structural features actually contain both semantic and structural information.

The structural concepts then can be learnt from one or multiple samples. In our model, the concept of one category is a distribution of position and relationship neurons for the category based on experiences. The sample distribution is used to approximated the prior distribution. The concepts will be further utilized in Block 6 to judge between possible candidates of recognition results.

\subsection{Integrated Recognition with Bayesian Learning (Block 5)}
Prior work has shown that perception can be interpreted as a Bayesian inference process from different pathways. Related model can predict human eye movements well in visual search tasks without any further assumptions or parameter tuning [45].

In this paper, the object recognition is considered as a Bayesian inference process based on models trained with different kinds of features. Firstly, recognition models, like softmax classifiers, are built based on different pathways, that is, different features, including episodic, semantic, and structural features. Each model outputs a vector of probabilities of all categories for an input sample. In training process, the correlations between features and categories are also learnt for feature-selection in Block 6. 

The recognition results are then inferred from the output probabilities of these recognition models by Bayesian learning. For computational convenience, this paper assumes that recognition based on different features are independent. The detailed computation process is as follows. 

\begin{eqnarray}
{\rm{P}}\left( {{{\rm{O}}_{\rm{i}}}{\rm{|}}{{\rm{M}}_1},{{\rm{M}}_2}, \ldots } \right) = \frac{{{\rm{P}}\left( {{{\rm{M}}_1},{{\rm{M}}_2}, \ldots {\rm{|}}{{\rm{O}}_{\rm{i}}}} \right){\rm{P}}\left( {{{\rm{O}}_{\rm{i}}}} \right)}}{{\mathop \sum \nolimits_j {\rm{P}}\left( {{{\rm{M}}_1},{{\rm{M}}_2}, \ldots {\rm{|}}{{\rm{O}}_{\rm{j}}}} \right){\rm{P}}\left( {{{\rm{O}}_{\rm{j}}}} \right)}}
\end{eqnarray}
\begin{eqnarray}
{\rm{P}}\left( {{{\rm{M}}_1},{{\rm{M}}_2}, \ldots {\rm{|}}{O_i}} \right)P\left( {{O_i}} \right) = P\left( {{O_i}} \right)\mathop \prod \limits_k P({M_k}|{O_i})
\end{eqnarray}
$O_i$ is the category of a certain object, ${M_i}$ are the output of recognition models based on different recognition features. For an object, the prior probabilities of each pathway are initialized as $\frac{{\rm{\varepsilon }}}{n}$, ${\rm{\varepsilon }}$ is a relative small number and n is the total number of pathways. During training, the prior distribution is updated by the sample distribution.

During recognition, post probabilities of potential categories are calculated and the category that maximizes the post probability is the output.

\begin{eqnarray}
{{\rm{O}}_{{\rm{output}}}} = \arg\max_i P{\rm{(}}{M_1},{M_2}, \ldots {\rm{|}}{O_i})
\end{eqnarray}

In short, by mimicking population coding and visual perception process, the proposed model can integrated different information extracted from original samples by Bayesian learning.

\subsection{Feature Re-selection (Block 6)}
\label{subsec:Feature} 

During recognition, human brain is not static but always adjusts and adapts dynamically to new stimuli. This paper especially focuses on the ambiguity of images. As the findings in visual systems suggest \cite{daelli2010recent}, for an ambiguous image which has multiple competitive candidates, human will pay more attention to the difference between the candidates. When a new category of images appears, the brain tends to form a new concept, based on existing semantic memory \cite{mcclelland2003parallel}

Inspired by the principles mentioned above, a feature re-selection strategy is applied to cope with ambiguous condition, in which the outputs of classification have more than one results with high confidence. The recognition process will then go backward to Block  $4$ to choose more distinctive structural features. For example, when the model cannot decide whether a handwritten digit is "$5$" or "$6$", it will go back to Block $4$ to choose the "horizon line" and "half circle" with a vertical relationship and focus on these features to distinguish between "$5$" and "$6$". These relationships between categories and features are learnt in Block $3$ and $4$.

The recognition models based on spatial positions and structural relationships are trained in Block $5$. The significance of different features is stored in the weights of those models. Given potential two candidate categories, the model will select features with more discriminative ability among two candidates, which is evaluated by the absolute differences of weights for the categories. If there are more than 2 candidates, the model averages the differences and use the mean as the significance of features. In this paper, to better utilize the features, we consider one block in the feature matrix (e.g. different positions of one semantic feature) as a whole. The corresponding weights of these features are summed up to get the total weight for the block. In re-selection process, the model automatically selects blocks with larger weights than average of all weights. 

\begin{figure}[!htb]
	\centering
	\subfigure[Structural relationship of digit "$6$"]
		{\includegraphics[width=0.45\linewidth]{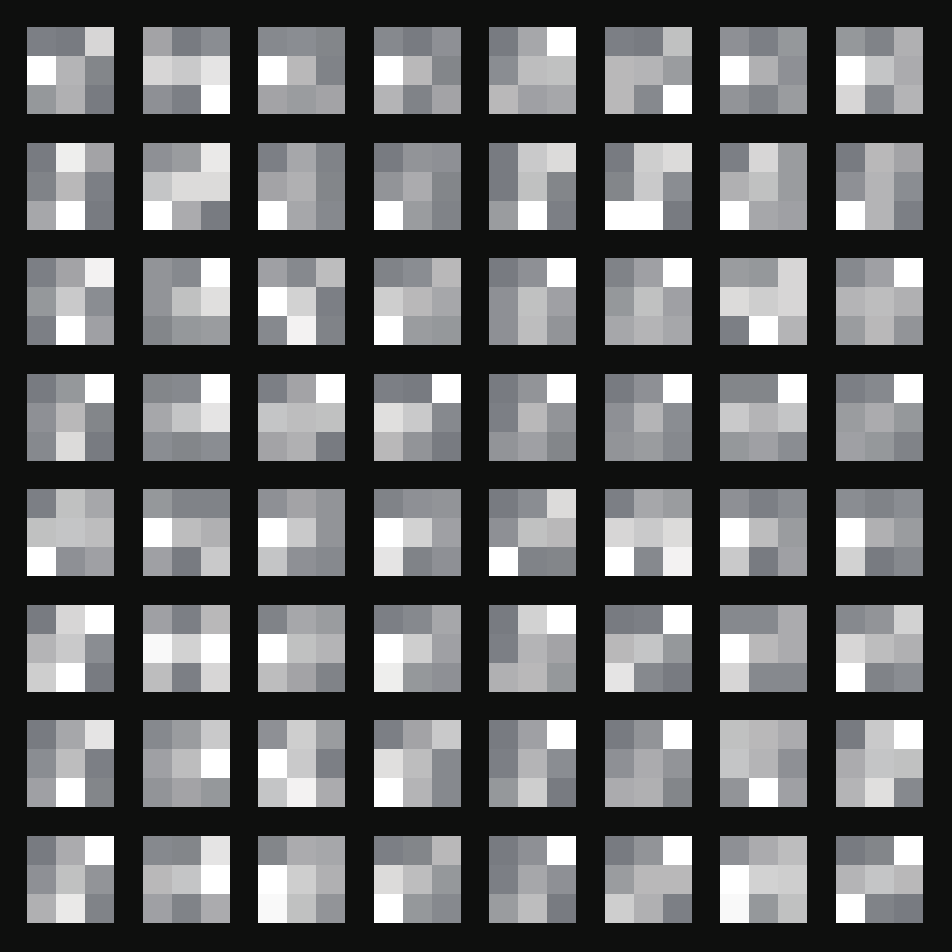} \label{fig:structural_relaiton_wo_selection}}
	\subfigure[Structural relationship of digit "$6$" after re-selection]
		{\includegraphics[width=0.45\linewidth]{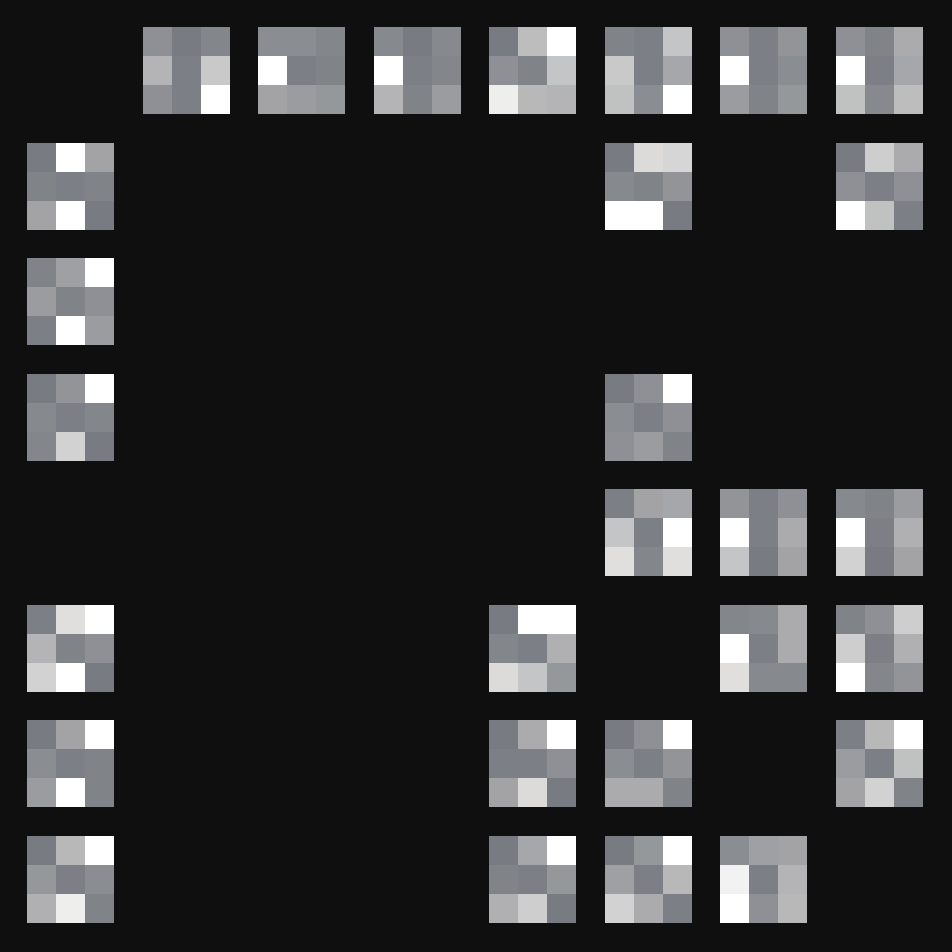}
			\label{fig:structural_relaiton_w_selection}}
	
	\caption{Structural concepts (relationships) and re-selected features of digit "$6$". The relationship matrix is built from all the samples in "$6$". The value of one cell represents the average activation on a certain PNeuron.}

	\label{fig:selected_feature}
\end{figure}

With the re-selected features, the judgment between "$5$" and "$6$"  is achieved by comparing structural information on the current sample and learnt concepts. The above mentioned concepts are the distribution of positions of semantic features and structural relationships.  As shown in Fig. \ref{fig:structural_relaiton_wo_selection}, structural concept of digit "$6$" is represented in a relationship matrix, which is generated by averaging all relationship matrices of category "$6$". It reveals how likely the neurons are activated when the input is "$6$". Fig. \ref{fig:structural_relaiton_w_selection} shows the selected significant features of "$6$" to distinguish "$6$" from "$5$". When an input image is ambiguous, only the activations of discriminative features would be feed to the classification models after re-selection. 

By applying the feature re-selection strategy, the proposed model achieves the ability of generalization and adaption to ambiguous input stimuli.

\section{EXPERIMENT}
Several experiments are conducted to verify the effectiveness of the proposed biologically inspired model, and each module is tested and analyzed in details. The experiments are focused on three aspects: 1) visualize the episodic and semantic features that are extracted by the proposed model; 2) investigate the structure information learned by the proposed model; 3) evaluate the classification performance on different datasets.

\subsection{Extraction of Episodic Features}

\begin{figure}[!htb]
	\centering
	\includegraphics[width=0.95\linewidth]{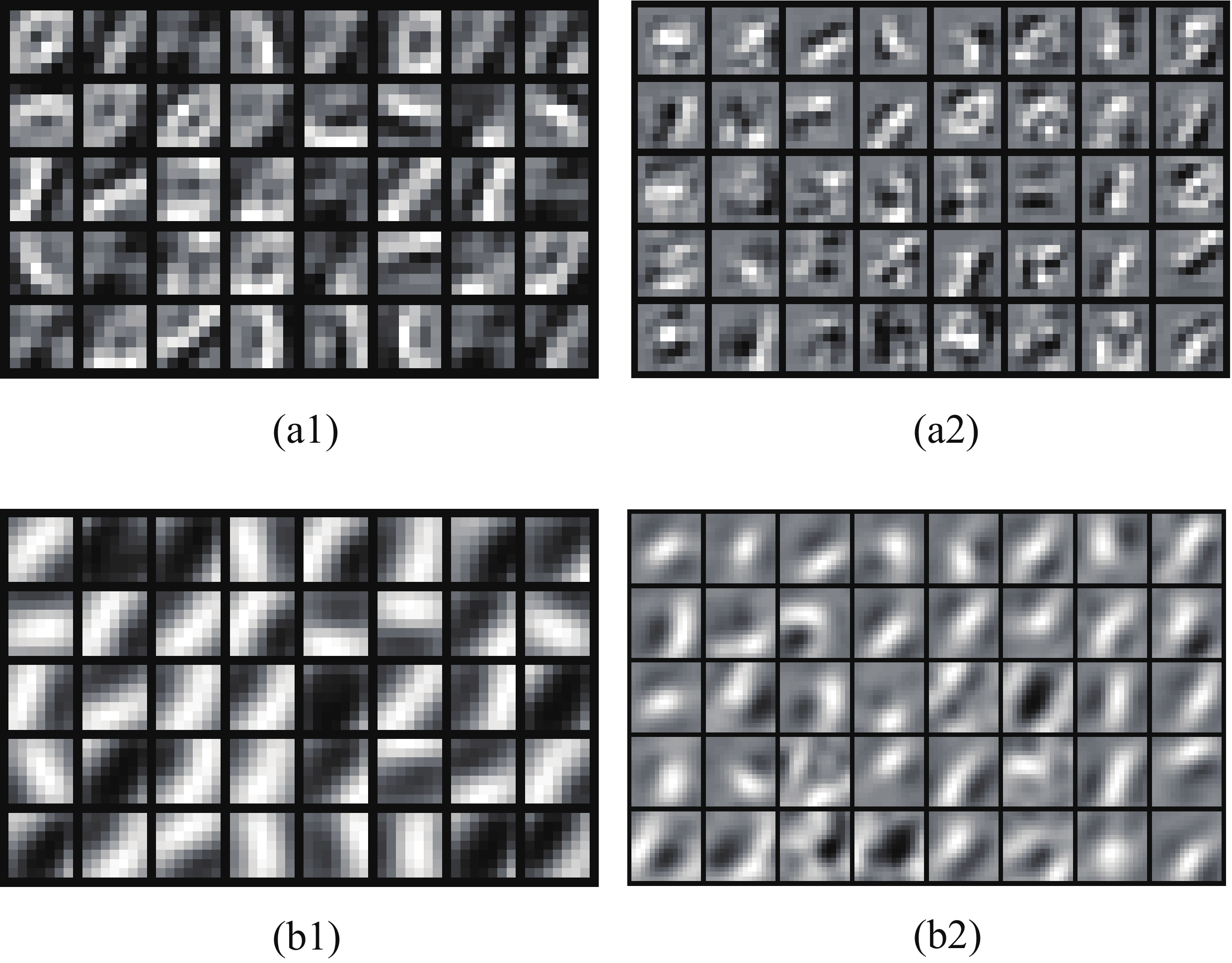}
	\caption{Illustration of episodic features extracted by the CDBN model. The learnt features in  (a) first layer and (b) second layer are the visualized, respectively. The upper row is visualized by deconvolution from  feature maps to the input space layer by layer, and the bottom row is visualized by averaging the top $100$ activations of inputs.}
	\label{fig: episodic_feature}
\end{figure}

This experiment is to visualize the extracted episodic features and to verify that these features can capture the critical information of the original image. Here, MNIST dataset is used as an example. The visualizations of the learnt weights of CDBN are given in Fig. \ref{fig: episodic_feature}, which corresponds to the episodic features in our model. Here, two visualization techniques are used, including the deconvolution method and the average of max activations used in \cite{lee2009convolutional}.

As is shown in Fig. \ref{fig: episodic_feature}, the proposed deconvolution method could achieve clearer edges and parts than the method in \cite{lee2009convolutional}. Furthermore, it is clear that the CRBM model can extract episodic features hierarchically from the original dataset. In details, features learnt by the first layer of the CRBM model are mostly edges and small details of the input digits, whereas the second layer of the CDBN model extracts more sophisticated components like circles and turning strokes. So it is reasonable to use the outputs of second layer to learn semantic features. From  Fig. \ref{fig: episodic_feature}, one may find out that some features are highly similar. One possible reason is that those features are trained without supervision, so some of them may more likely be attracted to the most significant features at the same time.

\begin{figure}[!htb]
	\centering
	\includegraphics[width=0.95\linewidth]{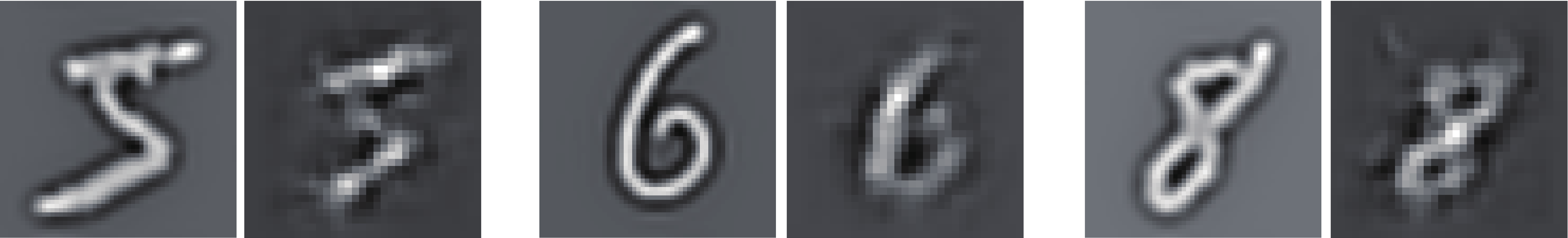}	
	\caption{Illustration of reconstruction from episodic features. Images on the left are the original patches, and the reconstruction images are shown on the right side.}
	\label{fig: reconstruction}
\end{figure}

To further verify that CDBN can extract and learn critical information of image, experiments on reconstruction from the episodic features are conducted. Some examples are given in Fig. \ref{fig: reconstruction}, which illustrates the high similarity between the original image and its reconstruction.

In addition, Fig. \ref{fig: reconstruction} illustrates how the learned weights encode episodic features. Original images can be directly reconstructed from high-level features, which indicates that most of the detailed information is captured by the second layer. 

\subsection{Semantic Features Extraction and Structure Learning}

This experiment is conducted to show the abstraction process from episodic features to structural semantic outputs.

\begin{figure}[!htb]
	\centering
	\subfigure[Deconvolution visualization]{\includegraphics[width=0.45\linewidth]{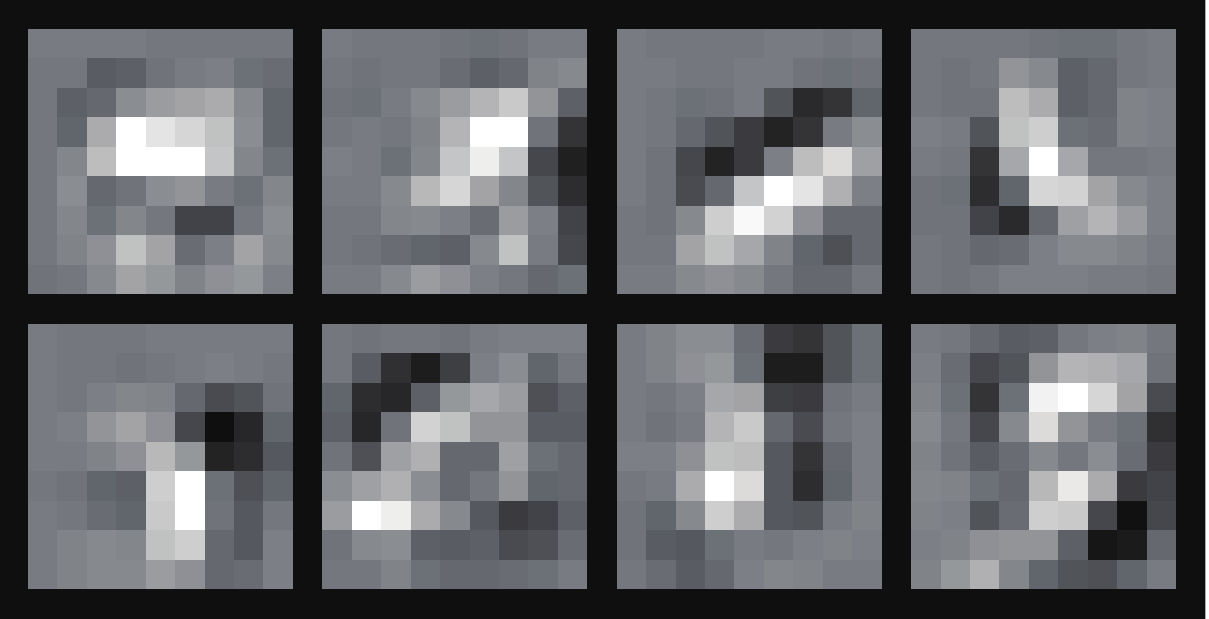}}
	\subfigure[Maximum Activation]{\includegraphics[width=0.45\linewidth]{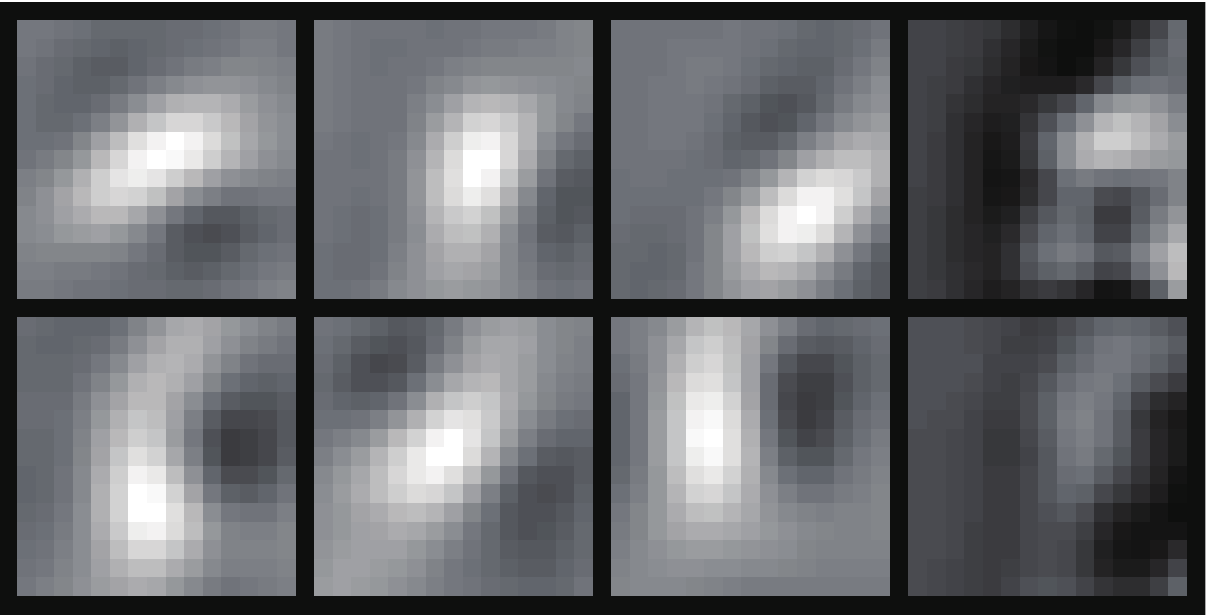}}	
	\caption{Illustration of semantic features visualized by (a) deconvolution and (b) maximum activation.}
	\label{fig: semantic_feature}
\end{figure}

Semantic features are clustered from extracted episodic features, as visualized in Fig. \ref{fig: semantic_feature}. The number of clusters is set as $8$. In Fig. \ref{fig: semantic_feature}, different semantic features are less similar with each other, which enhances the variety of features and captures more information with less features.

\begin{figure}[!htb]
	\begin{center}
		\includegraphics[width=0.5\linewidth]{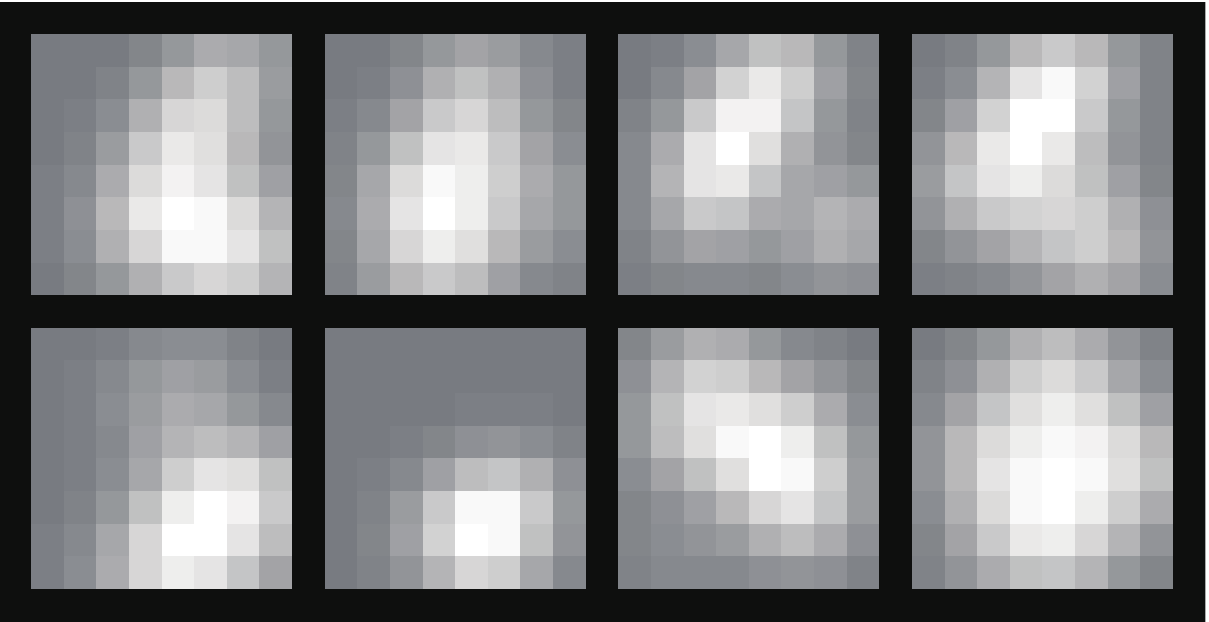}
	\end{center}
	\caption{Illustration of a position matrix. The distribution in one square matrix is like a mixture of Gaussian distributions.}
	\label{fig: position_matrix}
\end{figure}

After extracting semantic features, we then calculate the activations of PNeurons, by applying position tuning functions on the features. Here, a position tuning function is a 2D Gaussian function, with different mean but the same covariance matrix. For computational convenience, we use the discretized version of it, as illustrated in Fig. \ref{fig: position_matrix}. For each feature, there are $16$ PNeurons, which form a $4 \times 4$ position matrix. 

An example of position matrix is shown in Fig. \ref{fig: position_matrix}, which is similar with the mixture of several Gaussian distributions.  

Visualization of structural outputs is generated from the structural relationship matrix, which encodes the distributions and relative spatial relationships between features.

\begin{figure}[!htb]
	\begin{center}
		\includegraphics[width=0.7\linewidth]{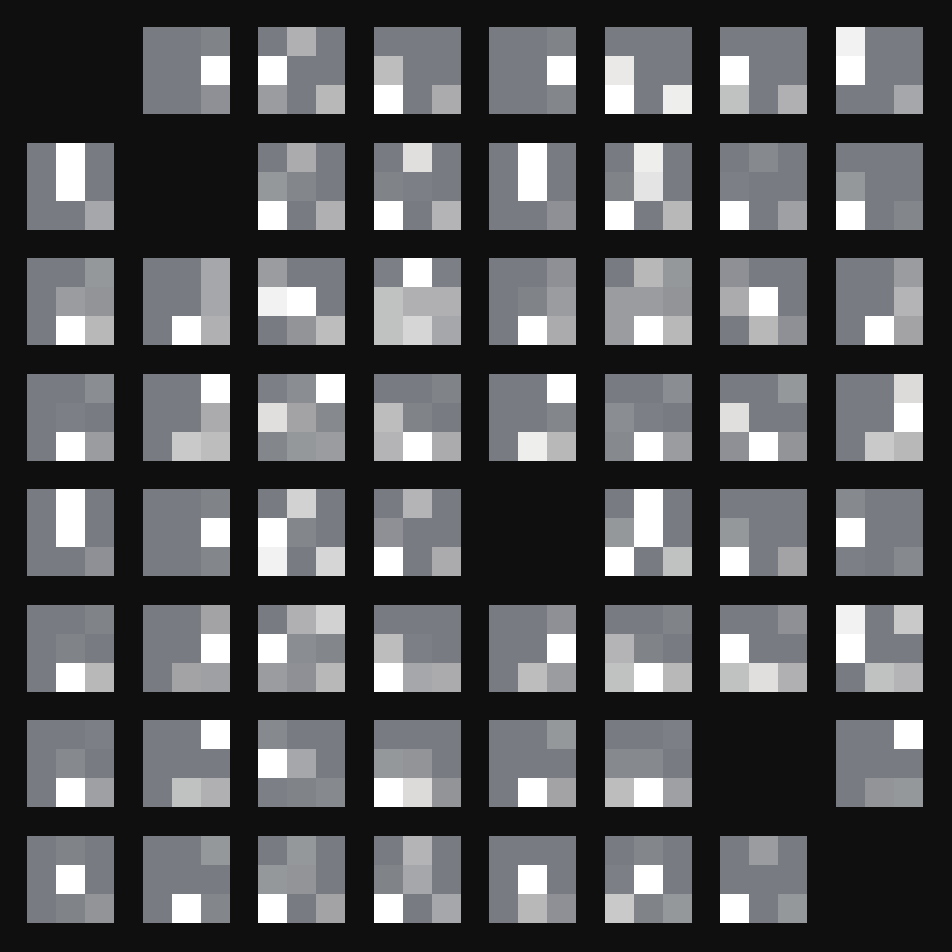}
	\end{center}
	\caption{Illustration of a structural relationship matrix. Each $3 \times 3$ square matrix represents the relationship between a pair of semantic features. The surrounding nodes encoded eight spatial relative directions between semantic features. The center nodes represent the spatial distance between the pair of semantic features.}
	\label{fig: structure_feature}
\end{figure}

Fig. \ref{fig: structure_feature} illustrates an example of structural relationship matrix, which is randomly selected from $100$ training samples of MNIST dataset. Each small square includes eight direction neurons (surrounding) and one distance neuron (center).

\subsection{Feature Re-selection Experiment}
\label{subsec:FeatureExp1}
The learnt semantic features and structures are not static during the testing process. Here we show how the features are re-selected to deal with the ambiguity and unfamiliarity.

When the input is ambiguous, it is easy to output as multiple candidates, the model could re-select features to achieve accurate classification. 

\begin{figure}[!htb]
	\centering
	\includegraphics[width=0.15\linewidth]{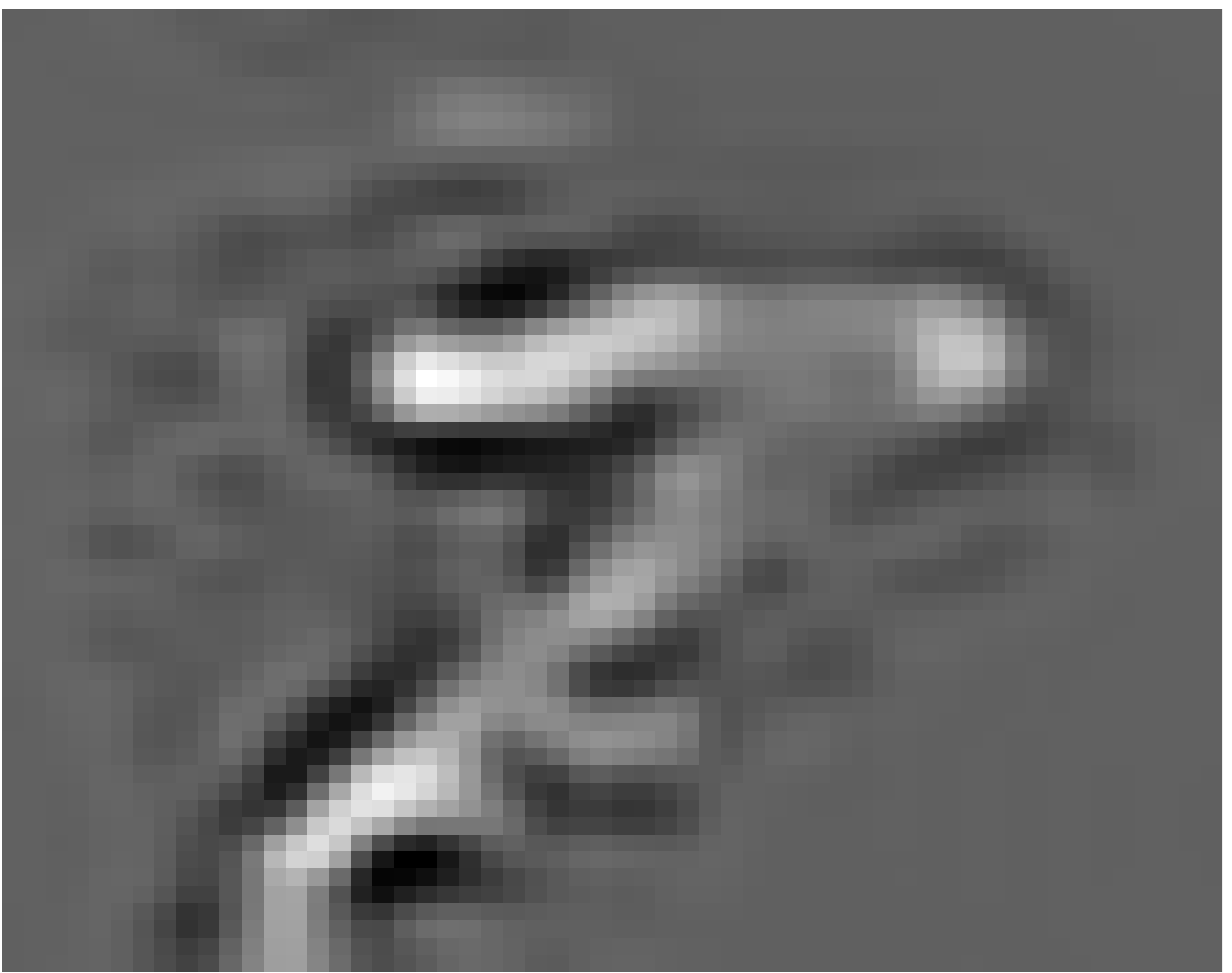}
	\includegraphics[width=0.15\linewidth]{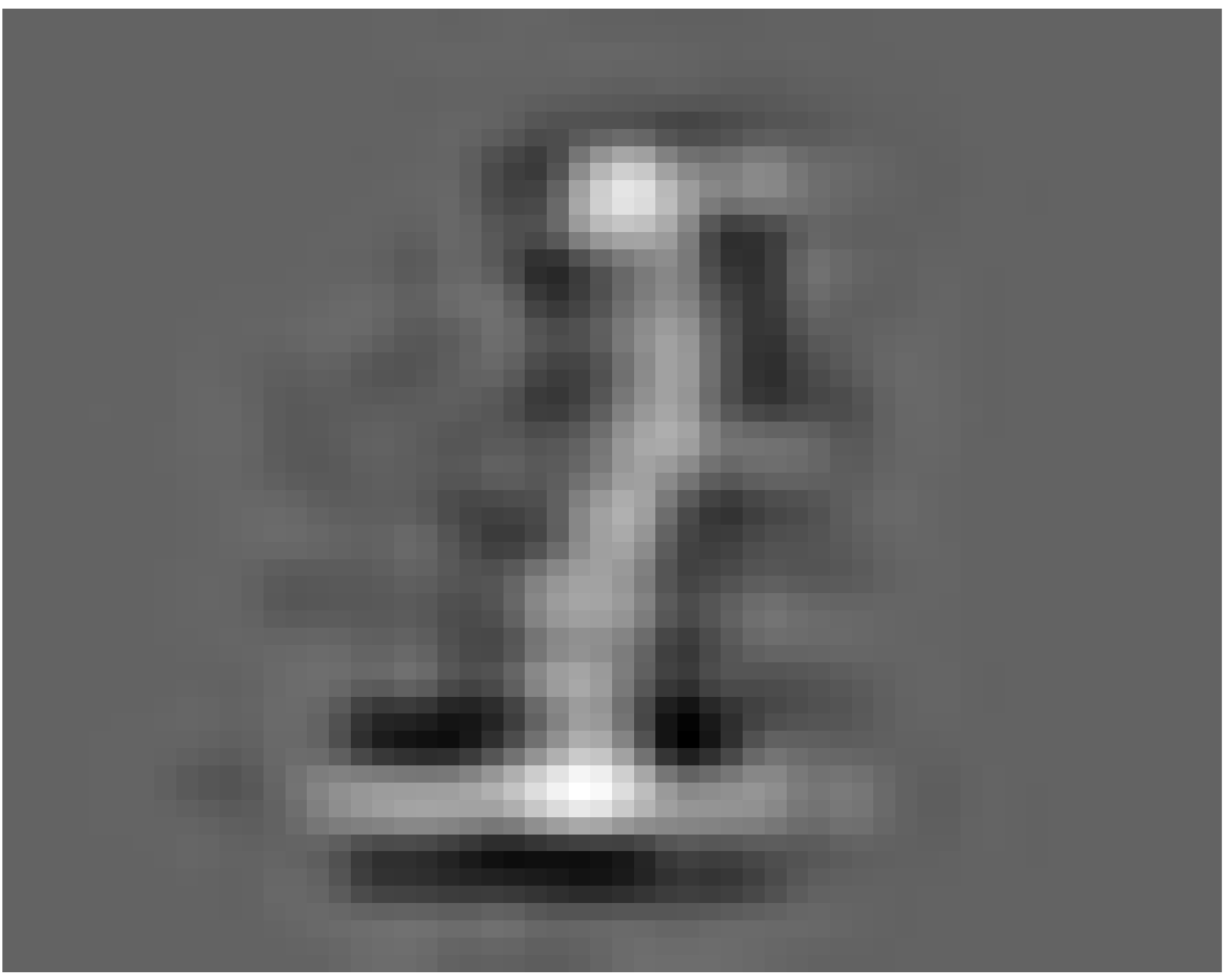}
	\includegraphics[width=0.15\linewidth]{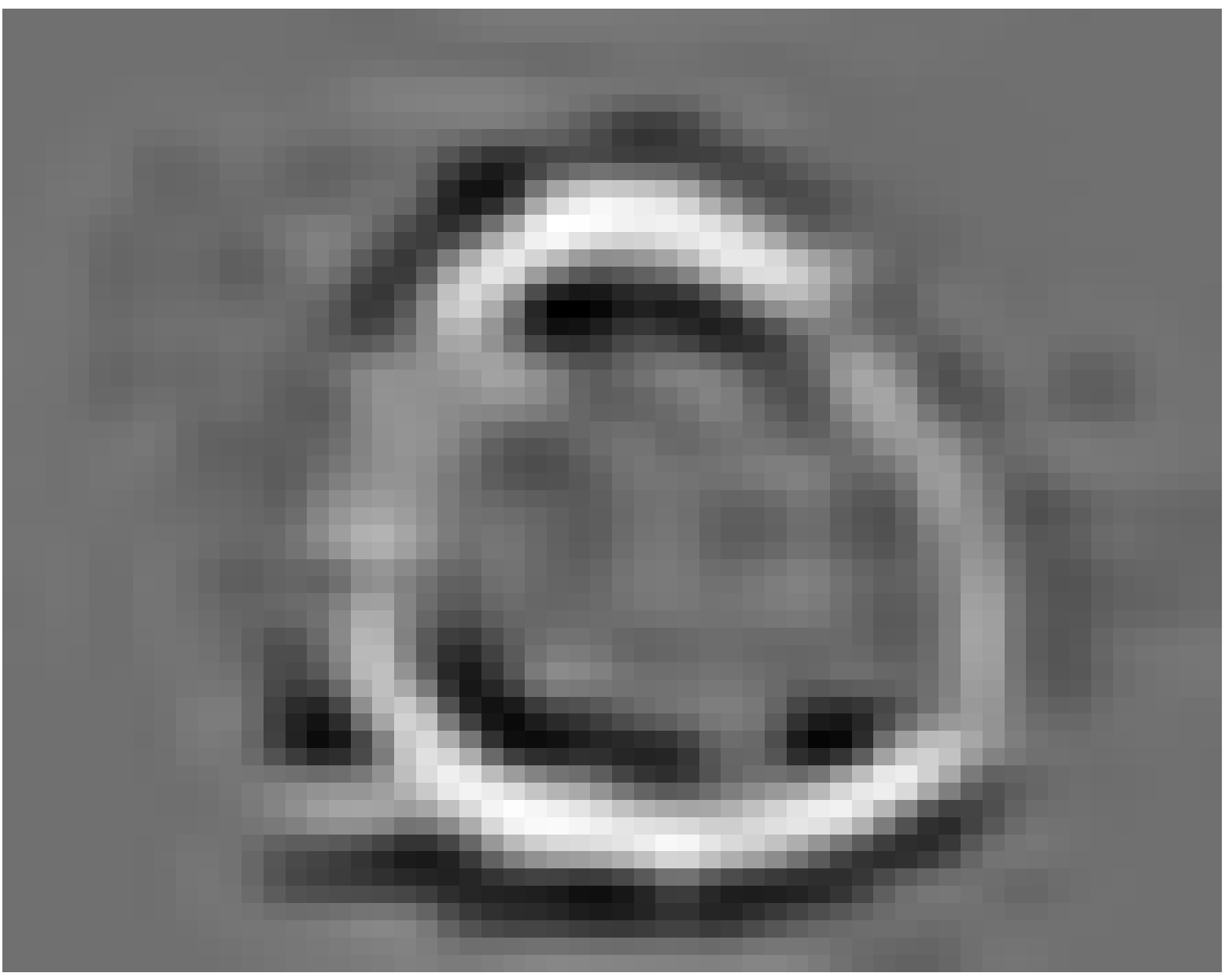}
	\includegraphics[width=0.15\linewidth]{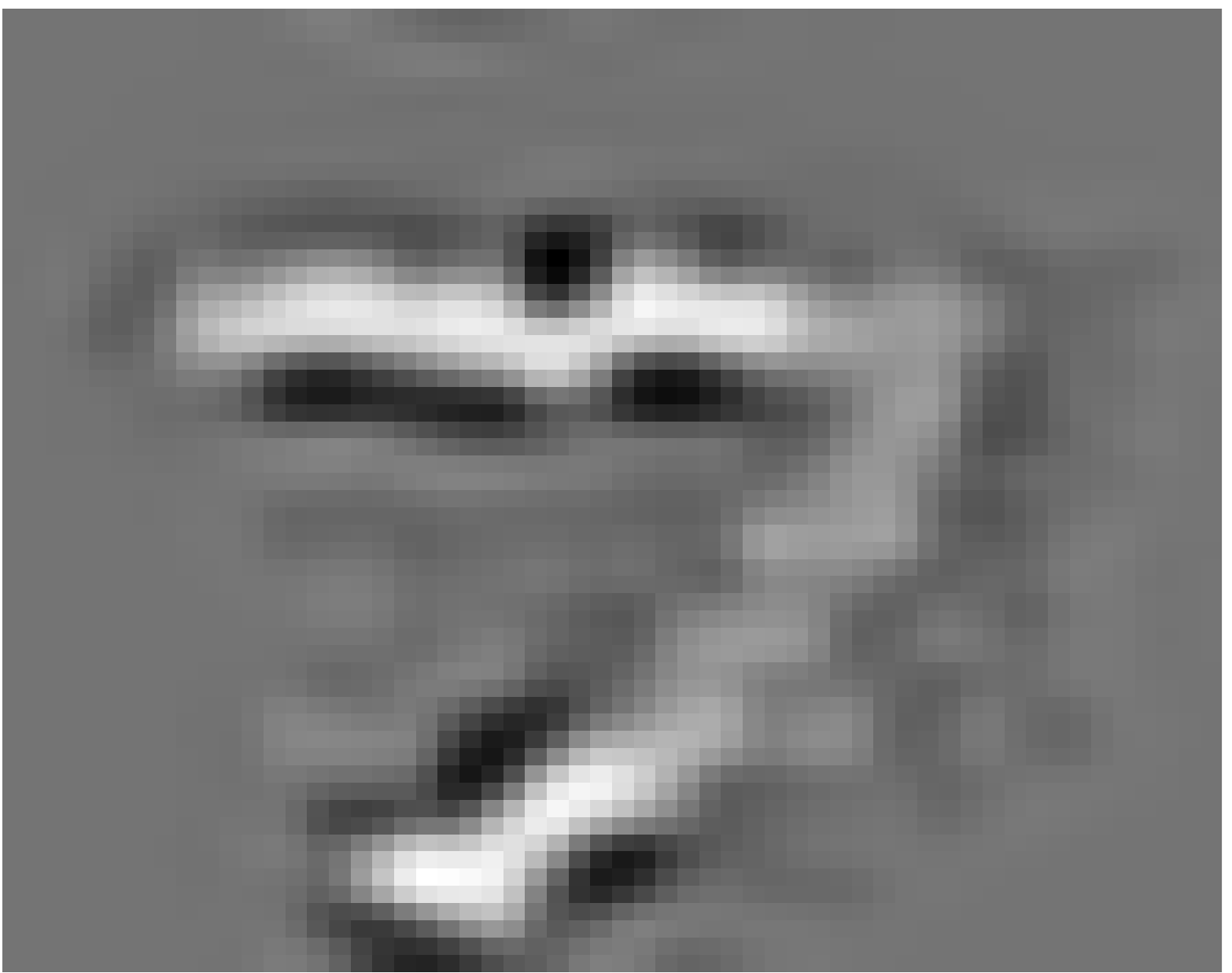}
	\includegraphics[width=0.15\linewidth]{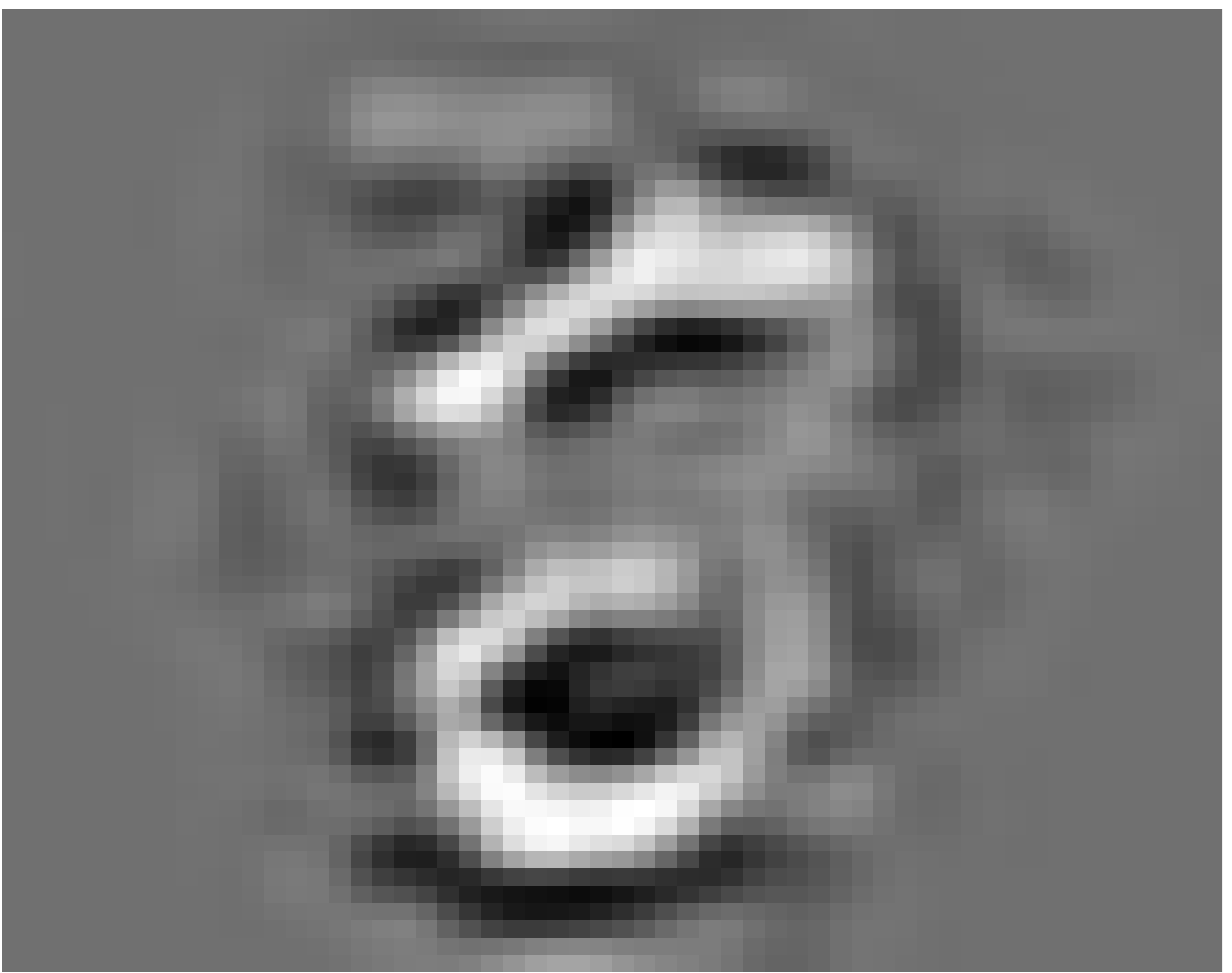}
	\caption{Examples of semantic ambiguous images. The images are optimized to be recognized as "$6$" by an ordinary convolutional neural network. Note that some image (like "$8$") is even ambiguous to human.}
	\label{fig:ambiguous_example}
\end{figure}

To illustrate the process in a better way, Fig. \ref{fig:ambiguous_example} shows examples of ambiguous images, whereas the input images are misclassified as "$6$" by a convolutional neural network. These ambiguous images are generated by the method proposed in \cite{goodfellow2014explaining}, so-called "adversarial images". That is, optimizing and modifying the original image, such as the one labeled as "$7$",  to be misclassified as "$6$" by a convolutional neural network. By applying the back-propagation to the input space and limiting the martingale of gradients, we are able to generate tiny perturbations to the original images which could mislead the model. Images with perturbation are originally designed for a convolutional neural network, but it could also affect the recognition by a CDBN, which is consistent with the results in \cite{goodfellow2014explaining}. 
\begin{figure}[!htb]
	\centering
	\subfigure[Position Matrix of "$5$"]
	{\includegraphics[width=0.45\linewidth]{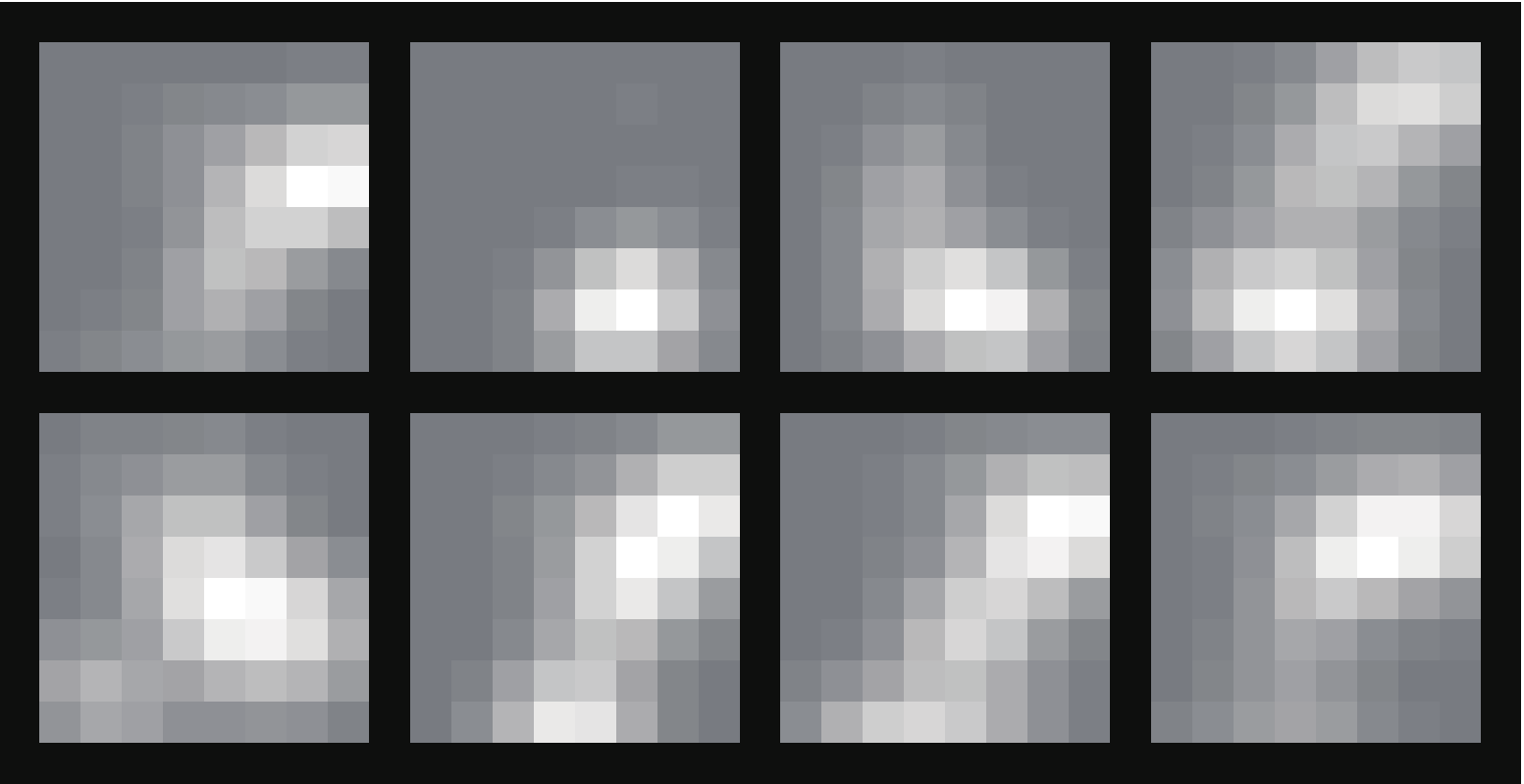} 
		\label{fig:position_matrix_5}}
	\subfigure[Position Matrix of "$6$"]
	{\includegraphics[width=0.45\linewidth]{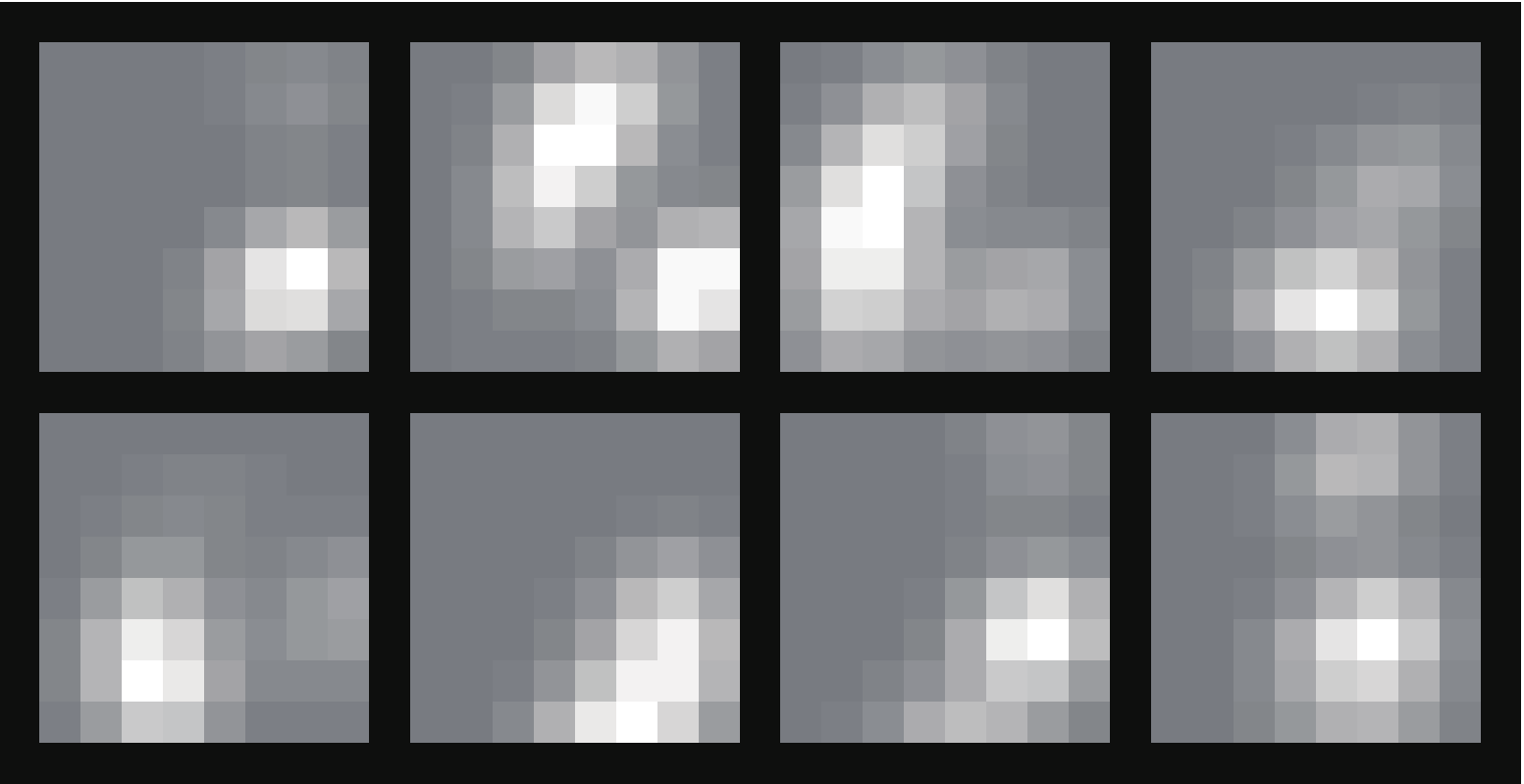} 
		\label{fig:position_matrix_6}}
	\subfigure[Structural Relationship of "$5$"]
	{\includegraphics[width=0.45\linewidth]{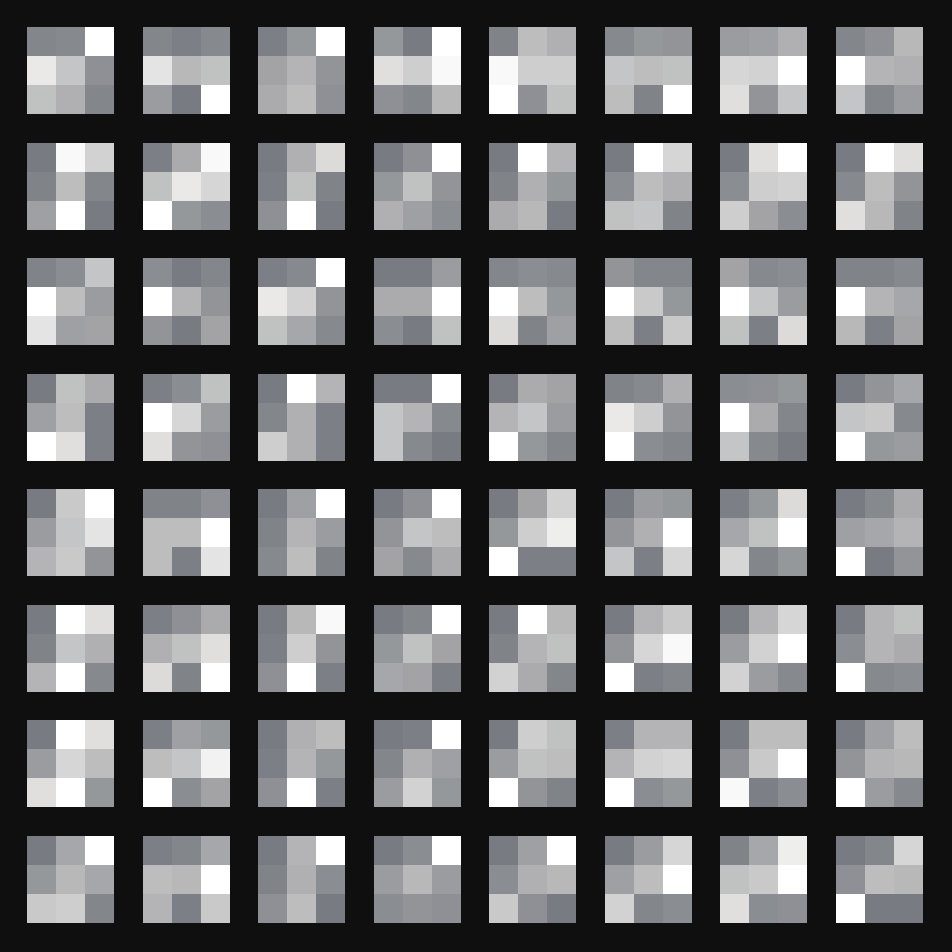} \label{fig:structural_relaiton_5}}
	\subfigure[Structural Relationship of "$6$"]
	{\includegraphics[width=0.45\linewidth]{RelationshipCenter6.eps}
		\label{fig:structural_relaiton_6}}
	
	\caption{Position matrix and structural relationships of category "$5$" and "$6$".}

	\label{fig:5_6_feature}
\end{figure}

\begin{figure}[!htb]
	\begin{center}
		\includegraphics[width=0.7\linewidth]{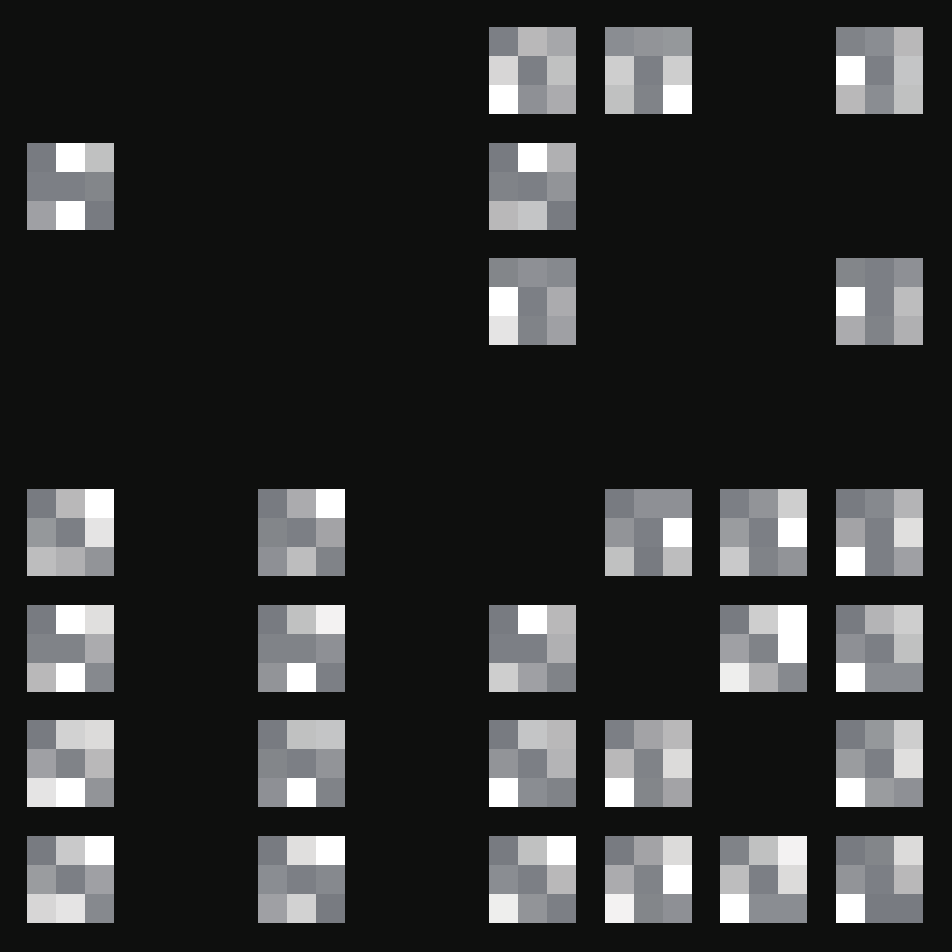}
	\end{center}
	\caption{Illustration of a structural relationship matrix after feature re-selection to distinguish "$5$" and "$6$".}
	\label{fig:select_feature}
\end{figure}

Following the strategy in Section \ref{subsec:Feature}, the significance of different features is learnt from training dataset. Fig. \ref{fig:5_6_feature} illustrates the learnt position matrix and structural relationships of digits "$5$" and "$6$". As is shown in Fig. \ref{fig:5_6_feature}, although "$5$" and "$6$" activate similar semantic features, the position and structural relationships between the features are quite different. Hence, by evaluating the differences of the position and structural relationship, we could find more distinctive features to build a new classifier, which is specific to separate "$5$" and "$6$". Fig. \ref{fig:select_feature} shows the chosen features after the re-selection to distinguish "$5$" from "$6$". 

\subsection{Classification Performance}
\subsubsection{MNIST with small training set}

The first comparison experiment is conducted between our model and some biologically inspired models (such as traditional CDBN\cite{lee2009convolutional}, HMAX\cite{riesenhuber1999hierarchical}, and etc.) on MNIST dataset. In total, MNIST includes $60000$ hand-writing digit images for training and $10000$ for testing. In this experiment, small training set is randomly and uniformly chosen from MNIST training data for ten categories. The code of HMAX model is obtained from the author's website. The traditional CDBN model has the same configuration and structure of our model, but without semantic features and structural information.

In this section, we use a two-layer CDBN with $40$ feature maps in the first layer and $40$ feature maps in the second layer. The pooling size is $2$ in both layers. The outputs of second layer are episodic features. From these episodic features, $8$ semantic features are extracted and then processed by the position tuning functions and structural relationship neurons. Two types of PNeurons are used with size of $16\ (4\times 4) $ and $64\ (8\times 8)$. For each pair of semantic features, we use 9 RNeurons (8 for direction, 1 for distance). 

The results are shown in Table $1$, performance of the proposed BSNN is better than HMAX and traditional CDBN. The main reason is because the semantic features are more discriminative even with a small number and integrated . Moreover, the performance is further improved by introducing the position neurons. 

\setcounter{table}{1}
\begin{table}[!htb] 
	\centering
	\textbf{{Table}\label{Tab:01}
		\arabic{table}.}{ Classification Error rate (\%) on MNIST \\(with $10$, $50$, $100$ training samples per category)}
	
	\begin{tabular}{|c|c|c|c|}
		\hline
		Samples per Class & $10$ & $50$ & $100$ \\
		\hline
		HMAX     &  $87.00$  &  $79.25$  &  $66.75$   \\
		\hline
		One Layer CDBN    & $35.75$ &  $30.25$  &  $16.25$   \\
		\hline
		Two Layer CDBN    & $24.50$ &  $16.25$  &  $12.35$  \\
		\hline
		Semantic Features    & $23.00$ &  $14.00$  &  $11.05$   \\
		\hline
		BSNN (16 PNeurons)   & $19.75$ &  $13.50$  &  $12.00$  \\ 	
		\hline
		BSNN (64 PNeurons)   & $19.50$  &  $12.00$ & $10.50$  \\	
		\hline
		BSNN Integrated & $\bm{17.25}$ & $\bm{9.75}$ & $\bm{7.29}$ \\
		\hline			
	\end{tabular}%
\end{table}%

\subsubsection{Ambiguous Images from MNIST}
The ambiguous data set is generated by adding a relative small perturbation to the original MNIST data sets \cite{cszegedy2014intriguing}. Details of the method are described in \ref{subsec:FeatureExp1}. Note that all the networks and classifiers are trained on the original MNIST data set. 

Table $2$ is the classification error rate of different models on the ambiguous images. Compared with the results in Table $1$, for HMAX and traditional CDBN, performances on ambiguous images are worse than those on the original images. After feature re-selection, the performance of BSNN has increased because the selected features are more discriminative than before.

\setcounter{table}{2}
\begin{table}[!htb] 
	\centering
	\label{Tab:02}
	\textbf{{Table}
		\arabic{table}.}{ Classification Error rate (\%) on Ambiguous MNIST \\( with $10$, $50$, $100$ training samples per category)}
	\begin{tabular}{|c|c|c|c|}
		\hline
		Samples per Class & $10$ & $50$ & $100$ \\
		\hline
		HMAX     &  $91.00$  &  $84.75$  &  $87.25$   \\
		\hline
		One Layer CDBN    & $53.00$ &  $43.75$  &  $23.25$   \\
		\hline
		Two Layer CDBN    & $44.75$ &  $35.50$  &  $20.00$  \\
		\hline
		Semantic Features    & $40.25$ &  $26.50$  &  $18.00$   \\
		\hline
		BSNN (16 PNeurons)   & $43.00$ &  $29.25$  &  $16.75$  \\ 	
		\hline
		BSNN (64 PNeurons)   & $35.75$  &  $23.25$ & $15.50$  \\	
		\hline
		BSNN Integrated & $30.00$ & $19.75$ & $\bm{13.00}$ \\
		\hline
		BSNN after Re-selection & $\bm{23.25}$ & $\bm{18.50}$ & $13.25$ \\
		\hline			
	\end{tabular}%
\end{table}%

\subsubsection{Facial shape dataset}
\begin{figure}[!htb]
	\centering
	\includegraphics[width=1\linewidth]{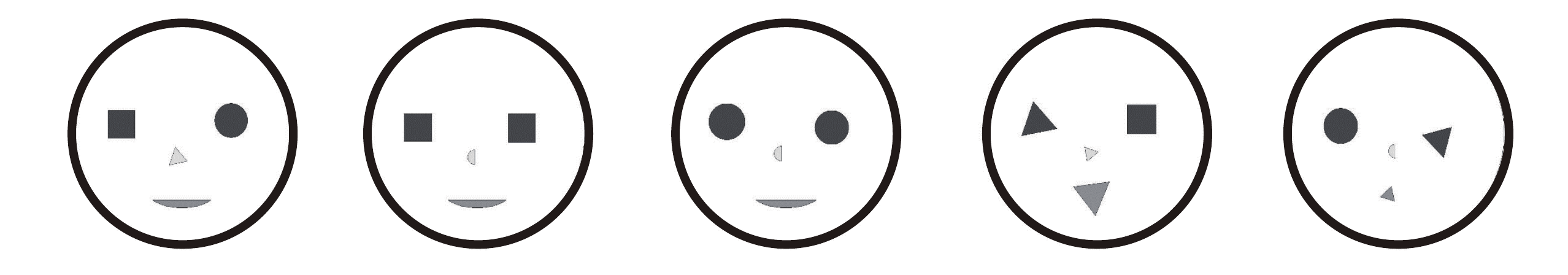}
	
	\caption{Artificial faces composed of different shapes. $5$ images are the representations of $5$ classes, respectively. The only difference between categories is the shape of each component.}
	\label{fig:facial_shape}
\end{figure}

The third comparison experiments are conducted on a facial shape dataset. The artificial face is composed of key components with different shapes. Some examples are shown in Fig. \ref{fig:facial_shape}. Compared to hand-written digits in MNIST, the facial shape dataset has more stable global structure, but also more scale- and shape-variant properties together with local transformations. 

\begin{figure}[!htb]
	\centering
	\includegraphics[width=1\linewidth]{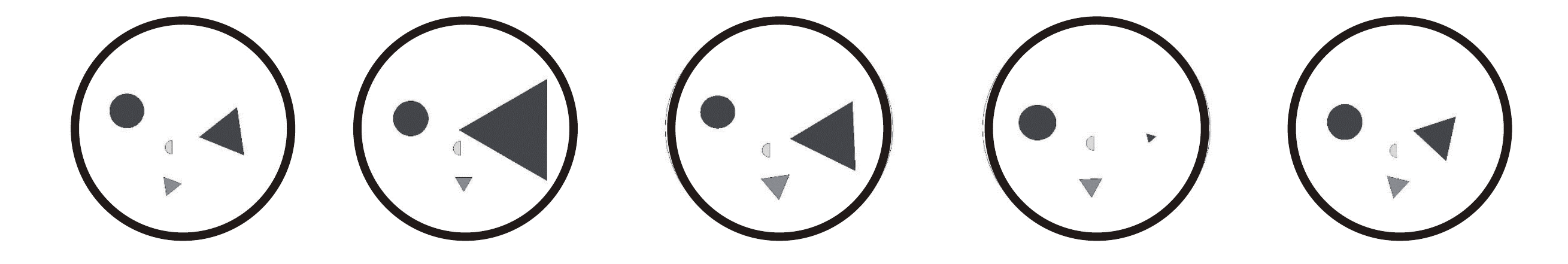}
	
	\caption{Illustration of artificial faces in one class (with circle left eye and triangle right eye). Although the shapes of key components are the same, the scales and locations of the components are various within a wide range.}
	\label{fig:facial_shape_one_class}
\end{figure}

More specifically, the dataset consists of $5$ classes of faces. As is shown in Fig. \ref{fig:facial_shape}, each face has four key components, including one mouth, one nose and two eyes. The main differences between classes are the different shapes of the components, where locations and scales of the components are also distributed in a wide range. Fig. \ref{fig:facial_shape_one_class} shows some examples in one class.


Conducting experiments as mentioned above, we compare the performance between HMAX, traditional CDBN and our model, with training data of different sizes ($30$, $100$, $300$ samples). The results are shown in Table $3$, which illustrate that the proposed model can successfully learn discriminative features even with a very small dataset.

\setcounter{table}{3}
\begin{table}[!htb] 
	\centering
	\label{Tab:02}
	\textbf{{Table}
		\arabic{table}.}{ Classification Error rate (\%) on Facial Shape Data \\( with $5$, $10$, $30$ training samples per category)}
	
	\begin{tabular}{|c|c|c|c|}
		\hline
		Samples per Category & $5$ & $10$ & $100$ \\
		\hline
		HMAX     &  $76.25$  &  $59.25$  &  $22.75$   \\
		\hline
		One Layer CDBN    & $37.16$ &  $17.50$  &  $0.5$   \\
		\hline
		Two Layer CDBN    & $16.25$ &  $4.50$  &  $0.00$  \\
		\hline
		Semantic Features    & $12.25$ &  $1.25$  &  $0.00$   \\
		\hline
		BSNN (16 PNeurons)   & $18.00$ &  $3.50$  &  $5.00$  \\ 	
		\hline
		BSNN (64 PNeurons)   & $8.00$  &  $3.00$ & $2.00$  \\	
		\hline
		BSNN Integrated & $\bm{7.00}$ & $\bm{0.50}$& $\bm{0.00}$ \\
		\hline				
	\end{tabular}%
\end{table}%

\section{Conclusion}
In this paper, a novel biologically inspired model is proposed for robust visual recognition, mimicking the visual processing system in human brain. By introducing semantics and structural conceptual outputs to the traditional CDBN network, the model gains more ability of generalization, especially for a small training dataset. The procedure of feature re-selection provides the model more robustness to ambiguity. During the cognition process, when ambiguity is detected during recognition process, new features according to the difference between ambiguous candidates are re-selected online for later cognition. 

In the future, the proposed model will be further improved by extracting spatiotemporal semantics and concepts for sequential analysis, which is more similar to human neural system. Another approach to enhance the model is to further introduce the biological mechanisms in higher level perception and inference. A more flexible and robust classifier, such as the function of prefrontal cortex in human, is useful to process different outputs in an integrated manner. 

\bibliographystyle{IEEEtran}
\bibliography{IEEEabrv,reference}

\end{document}